\documentclass{article}

\usepackage{arxiv}

\usepackage[utf8]{inputenc} % allow utf-8 input
\usepackage[T1]{fontenc}    % use 8-bit T1 fonts
\usepackage{hyperref}       % hyperlinks
\usepackage{url}            % simple URL typesetting
\usepackage{booktabs}       % professional-quality tables
\usepackage{amsfonts}       % blackboard math symbols
\usepackage{nicefrac}       % compact symbols for 1/2, etc.
\usepackage{microtype}      % microtypography
\usepackage{xcolor}         % colors

\usepackage{amsmath}
\usepackage{amsthm}
\usepackage{algorithm}
\usepackage{algpseudocode}
\usepackage{enumitem}
\usepackage{graphicx}
\usepackage{subcaption}
\usepackage{caption}

\newtheorem{case}{Case}

\newtheorem{definition}{Definition}

\newtheorem{theorem}{Theorem}

\newtheorem{lemma}{Lemma}

\newcommand{\bbox}{\hfill $\Box$}

\newcommand \ETC {ETC-TPZSG}
\newcommand \ETCAE {ETC-TPZSG-AE}
\newcommand \ETCNUE {ETC-TPZSG-AE-NUE}

\DeclareMathOperator*{\argmax}{argmax}
\DeclareMathOperator*{\argmin}{argmin}

\title{Two-Player Zero-Sum Games with Bandit Feedback}

\author{ Elif Yılmaz \\ University of Neuchâtel \\ Neuchâtel, Switzerland \\ \texttt{elif.yilmaz@unine.ch}  \And Christos Dimitrakakis \\ University of Neuchâtel \\ Neuchâtel, Switzerland \\ \texttt{christos.dimitrakakis@unine.ch}   }

\begin{document}
\maketitle

%%% Use this environment to specify a short abstract for your paper.

\begin{abstract}
  We study a two-player zero-sum game in which the row player aims to maximize their payoff against a competing column player, under an unknown payoff matrix estimated through bandit feedback. We propose three algorithms based on the Explore-Then-Commit (ETC)  and action pair elimination frameworks. The first adapts it to zero-sum games, the second incorporates adaptive elimination that leverages the $\varepsilon$-Nash Equilibrium property to efficiently select the optimal action pair, and the third extends the elimination algorithm by employing non-uniform exploration. Our objective is to demonstrate the applicability of ETC and action pair elimination algorithms in a zero-sum game setting by focusing on learning pure strategy Nash Equilibria. A key contribution of our work is a derivation of instance-dependent upper bounds on the expected regret of our proposed algorithms, which has received limited attention in the literature on zero-sum games. Particularly, after $T$ rounds, we achieve an instance-dependent regret upper bounds of $O(\Delta + \sqrt{T})$ for ETC in zero-sum game setting and $O\left(\frac{\log (T \Delta^2)}{\Delta}\right)$ for the adaptive elimination algorithm and its variant with non-uniform exploration, where $\Delta$ denotes the suboptimality gap. Therefore, our results indicate that the ETC and action pair elimination algorithms perform effectively in zero-sum game settings, achieving regret bounds comparable to existing methods while providing insight through instance-dependent analysis.
\end{abstract}

\section{Introduction}
This paper focuses on finite two-player zero-sum games (TPZSGs), where two players interact in a setting with strictly opposing objectives, as first analyzed in \cite{v1928theorie, vonneumanandmorgenstern}. These games play a foundational role in game theory and online learning, capturing adversarial interactions in various applications such as economic competition and adversarial machine learning. In such settings, players repeatedly select actions from finite sets, with payoffs determined by a fixed but unknown game matrix. 

Studies on learning in games typically use either \emph{full information feedback} where players observe the opponent’s actions or the entire payoff matrix after each round or \emph{partial feedback}, also called  bandit feedback, where each player observes only their own payoffs without access to the opponent’s strategy or the full matrix \cite{blum2007external}. Zero-sum games with bandit feedback setting, where players also observe each other's actions, were introduced in \cite{o2021matrix}. In this paper, we study a TPZSG setting under bandit feedback with centralized control over both players by learning pure strategies through balancing of exploration and exploitation.

Several existing studies have advanced our understanding of how players learn to play optimally in zero-sum games with bandit feedback. The variant of UCB algorithm for zero-sum games under bandit feedback has been analyzed using worst-case regret bounds in \cite{o2021matrix}. However, worst-case ignores the specific structure of the game. Instance-dependent analysis adapts the regret guarantees to the specific properties of a game to provide more practical performance guarantees. More recently, instance-dependent regret bounds have been investigated in TPZSGs \cite{ito2025instance, maiti2025limitations}. However, an understanding of instance-dependent regret in TPZSGs under bandit feedback remains incomplete.

\paragraph{Contributions.} This paper investigates the effectiveness of the \emph{Explore-Then-Commit} (ETC) and action elimination algorithms in TPZSGs. While ETC has been widely studied in standard multi-armed bandit problems, there has been no prior work specifically analyzing its performance to zero-sum game settings.  Thus, our work is the first to provide a detailed analysis of ETC in this context. A key motivation for studying ETC is its algorithmic simplicity, which enables a clear structure. Our primary objective is to determine whether ETC and action elimination algorithms are applicable to zero-sum games by offering a refined understanding of regret behavior.

In particular, we propose three algorithms that adapt the ETC and action pair elimination approaches to the TPZSG setting and provide their instance-dependent regret analyses. The first one follows the classical ETC framework by uniformly selecting actions for a fixed number of steps and then committing to a pure Nash Equilibria (NE), achieving a regret upper bound of  $O(\Delta + \sqrt{T})$ (Section \ref{section:ETC_and_regret}). The second algorithm incorporates adaptive elimination (AE), similar to \cite{auer2010ucb}, by sequentially eliminating actions that, with high probability, are not part of an $\varepsilon$-Nash Equilibrium; this yields an upper regret bound of $O\left(\frac{\log (T \Delta^2)}{\Delta}\right)$ (Section \ref{section:elimination_and_regret}). Intuitively, this approach is expected to converge more rapidly to the optimal strategy by efficiently reducing uncertainty and narrowing the action sets of the players. Finally, the third algorithm extends the adaptive elimination approach through non-uniform exploration, also achieving the same regret bound (see Section \ref{section:nonuniform_exploration}). Consequently, our results demonstrate that ETC and elimination approaches can be highly effective in zero-sum game settings.

%\paragraph{Paper structure.} Section \ref{section:related_work} reviews related work. Section \ref{section:preliminaries} introduces preliminaries and formalizes a two-player zero-sum game setting. In Section \ref{section:ETC_and_regret}, we present the ETC in a TPZSG setting and analyze its regret. Section \ref{section:elimination_and_regret} introduces an adaptive elimination algorithm and provides its regret analysis. In Section \ref{section:nonuniform_exploration}, we extend the elimination algorithm by using the non-uniform exploration approach and analyze its regret upper bound. Section \ref{section:experiments} offers empirical results to validate their theoretical performances. Finally, Section \ref{section:discussion} discusses our conclusions and future research directions.

\section{Related Work}
\label{section:related_work}
The multi-armed bandit (MAB) problem was introduced by \cite{thompson1933likelihood} and originally described by \cite{robbins1952some}. It has since become a crucial framework for online decision-making under uncertainty, as explored in works such as \cite{audibert2009exploration, auer2002finite, auer1995gambling, auer2010ucb, blum2007learning, bubeck2012regret}. The MAB problem models scenarios in which a player must repeatedly choose from a set of actions to maximize their own rewards while balancing exploration-exploitation tradeoff. This formalization was established by \cite{lai1985asymptotically} and further investigated in studies such as \cite{audibert2007tuning, maes2012learning}. 

A central challenge is identifying the action with the highest expected reward while minimizing suboptimal choices, which require efficient exploration and statistical confidence. Several approaches guarantee low error probabilities within finite trials, in order to minimize regret \cite{audibert2010best, gabillon2012best}. Since the player has no prior knowledge of the expected rewards, a dedicated exploration phase is essential, which has motivated numerous studies such as \cite{bubeck2009pure, neu2015explore}.

The Explore-Then-Commit (ETC) algorithm, which has been introduced by \cite{perchet2016batched}, is one of the fundamental approaches in the MAB setting. It explores each action a fixed number of times, then commits to the best performing action for the remaining rounds. Due to its simplicity and effectiveness, variants of the ETC algorithm are widely used in various decision-making scenarios \cite{garivier2016explore, yekkehkhany2019etc}. 

Zero-sum games represent a core concept in game theory, which describe competitive scenarios in which the gain of one player is exactly equal to the loss of the other. First highlighted by \cite{v1928theorie, vonneumanandmorgenstern}, their theory and applications have been extensively explored \cite{blum2007learning, cesa2006prediction, washburn2014two}. The Minimax Theorem in \cite{v1928theorie, vonneumanandmorgenstern} guarantees the existence of equilibrium in TPZSGs, but finding equilibrium becomes significantly harder when the payoff matrix is unknown \cite{daskalakis2011near, daskalakis2009complexity, mazumdar2019finding, zinkevich2007new}.

Furthermore, the concept of Nash Equilibrium, which is introduced by \cite{nash1950equilibrium, nash1951non}, describes optimal strategies of players in which no player is incentive to deviate. The study \cite{daskalakis2009complexity} shows that converging to such an equilibrium is challenging. The concept of $\varepsilon$-Nash Equilibrium provides an approximate solution where the strategy of each player is within a tolerance level $\varepsilon$. \cite{daskalakis2009note} uses this property to efficiently approximate equilibrium in two-player games.

In a zero-sum game setting with bandit feedback, players engage in repeated interactions without knowledge of the true payoff matrix. Various approaches, such as UCB \cite{o2021matrix}, Tsallis-INF \cite{ito2025instance} and a novel algorithm \cite{maiti2025limitations}, have been proposed to efficiently estimate the unknown payoff matrix while minimizing regret. Learning the game matrix is crucial for converging to equilibrium strategies and ensuring optimal decision-making in such environments.  In addition, \cite{maiti2023instance} and \cite{maiti2024near} investigate instance-dependent sample complexity bounds while \cite{cai2023uncoupled} examines the convergence rates to NE. 

While \cite{o2021matrix} focuses on the worst-case regret analysis, achieving $O(\sqrt{T})$, and \cite{maiti2025limitations} analyzes instance-dependent regret only for a $2\times2$ game matrix, \cite{ito2025instance} provides an instance-dependent analysis achieving an upper bound of $O \left( \left(\sum_{i \neq i^*} \frac{1}{\Delta_i} + \sum_{j \neq j^*} \frac{1}{\Delta'_j}\right) \log T \right)$ in the case of pure equilibria where $\Delta$ and $\Delta'$ denote some suboptimality gaps. Consequently, developing a comprehensive understanding of instance-dependent regret in TPZSGs under bandit feedback remains insufficiently addressed. In contrast to \cite{ito2025instance}, where both players independently follow the Tsallis-INF algorithm, our setting is closely related to \cite{o2021matrix} and \cite{maiti2025limitations}, but we assume control over both players. Our goal is to demonstrate the applicability of ETC and action elimination algorithms to zero-sum games and to achieve comparable regret bounds.

\section{Preliminaries}
\label{section:preliminaries}
In this section, we define the basic concept of a TPZSG such as payoff matrix, the regret definitions we focus on, and the gaps $\Delta$ that we use for the analysis. In a TPZSG, the goals of the players are strictly opposed, which implies that any gain by one player corresponds to an equal loss by the other. The outcome of such a game is fully determined by the strategies chosen by players and the goal of each player is to maximize their own payoff while minimizing that of their opponent.

%\paragraph{Notation.} Throughout this paper, $O(\cdot)$ denotes an upper bound up to constant factors.

\subsection{Two-Player Zero-Sum Games}
In a two-player zero-sum game, consisting of a row $(x)$ player  and a column $(y)$ player, the row player aims to maximize their own payoff while the column player attempts to take an action to minimize this payoff.   

Let $S_x$ and $S_y$ denote the action sets of the row and column players, respectively, with the joint action space $S_A = S_x \times S_y$. We define $m = |S_x|$ and $l = |S_y|$  as the numbers of actions available to the row and column players, and let $N = ml = |S_A|$ be the total number of action pairs. The game is specified by a payoff matrix $A$; therefore, if $(i, j)$ is played, the row player receives a payoff of $A(i,j)$ and the column player receives $-A(i, j)$. Moreover, the row player has a strategy (i.e. a probability distribution) $p$ to select an action $i \in S_x$, while the column player selects an action $j \in S_y$ according to a strategy $q$.

\begin{definition}
\label{def:mixed_strategy}
A pair of mixed strategies $(p^*, q^*)$ is a Nash Equilibrium if for all strategies $p$ and $q$, it satisfies
\begin{align} 
    {p}^\top A q^* \leq {p^*}^\top A q^*  \leq {p^*}^\top A q
\label{eq:minmaxgame} 
\end{align} 
where ${p^*}^\top A q^*$ is the value of the game \cite{nash1950equilibrium}. 
%This means that $p^*$ and $q^*$ are optimal strategies for row and column players, respectively. 
\end{definition}

There always exists an optimal mixed strategy for both players that guaranties the best possible outcome according to the Minimax Theorem \cite{v1928theorie, vonneumanandmorgenstern}. When a mixed strategy assigns a probability of one to a single action and zero to all others, it refers to a \emph{pure} strategy. In our setting, we focus on a TPZSG where both players adopt pure strategies.

\begin{definition}[Pure Nash Equilibria]
\label{def:pure_strategy}
A pair $(i^*, j^*)$ is a unique pure NE if $i^* = \argmax_{i \in S_x} \min_{j \in S_y} A (i,j)$ and $j^* = \argmin_{j \in S_y} \max_{i \in S_x} A (i,j)$ or equivalently
\begin{align}
     A(i,j^*) \leq A(i^*,j^*) \leq A(i^*,j),  \quad \forall i \in S_x, j \in S_y.
     \label{eq:nasheqproperty}
\end{align}
\end{definition}
The condition in \eqref{eq:nasheqproperty} ensures that no player has an incentive to deviate to another action.

\subsection{Games with Bandit Feedback}
The game proceeds in rounds. At time $t$, the row player draws an action $i_t \in S_x$ from a strategy $p_t$, and the column player draws $j_t \in S_y$ from $q_t$. The players then observe a noisy payoff $r_t$ with $\mathbb{E}[r_t \mid i_t, j_t] = A(i_t, j_t)$, where the maximizing player receives $r_t$ and the minimizing player receives $-r_t$. We assume that both players observe each other's action. \cite{o2021matrix} introduced this setting with algorithms for row player against any column player; we instead assume centralized control over both players.

Since the players do not know $A$, we use the notation $\hat{A}(i,j)$ to denote the estimated payoff when the action pair $(i,j)$ is played. In each round $t$, the estimated game matrix is updated by computing the average payoff for each action pair $(i,j)$ as follows: 
\begin{align}
\hat{A}_t(i,j) = \frac{1}{n_{ij,t}} \sum_{s=1}^t \mathbb{I} \{ (i_s,j_s) = (i,j) \} r_s  
\label{eq:averagepayoff}
\end{align}
where $n_{ij,t}$ is the number of times the action pair $(i,j)$ played until round $t$, $r_s$'s are independently and identically distributed (iid) $\sigma$-subgaussian payoffs. 

\subsection{Gap Notions}
To quantify how close a chosen action pair is to the NE, we define suboptimality gaps based on deviations from the equilibrium and the best responses of the players. These gaps capture the extent to which each player could improve their outcome by deviating from their current strategy. For any action $i \in S_x, j \in S_y$, we define the following suboptimality gaps:
\begin{align}
    & \Delta_{ij}^\text{max} = \max_{i' \in S_x} A(i',j) - A(i,j) \\
    & \Delta_{ij}^\text{min} = A(i,j) - \min_{j' \in S_y} A(i,j') \\
    & \Delta_{ij} =  \Delta_{ij}^\text{max} +  \Delta_{ij}^\text{min}  \\ 
    & \Delta_{ij}^* = A(i^*,j^*) - A(i,j)
\end{align}
We note that $\Delta_{ij}^\text{max} \geq 0$ and $\Delta_{ij}^\text{min} \geq 0$, therefore $\Delta_{ij} \geq 0$ for any action pair $(i,j)$. However, $\Delta_{ij}^*$ might be negative. 

\begin{lemma}
\label{lemma:relationbetw_delta*_and_delta}
    For all $i \in S_x$ and $j \in S_y$, $\Delta_{ij}^* \leq \Delta_{ij}$.
\end{lemma}
\begin{proof}
    Using the NE property in \eqref{eq:nasheqproperty}, for all $i \in S_x$ and $j \in S_y$ we can write $A(i^*,j^*)  \leq A(i^*,j) \leq \max_{i' \in S_x} A(i', j)$, which implies
    \begin{align}
        & A(i^*,j^*) - A(i,j)  \leq \max_{i' \in S_x} A(i', j) - A(i,j).
        \label{ineq:relation_delta*_delta_max}
    \end{align}
    Similary, since $A(i^*,j^*)  \geq A(i,j^*) \geq \min_{j' \in S_y} A(i, j')$ we have the following:
    \begin{align}
        & A(i^*,j^*) - A(i,j) \geq \min_{j' \in S_y} A(i, j') - A(i,j). \label{ineq:relation_delta*_delta_min}
    \end{align}
    Then, the inequality \eqref{ineq:relation_delta*_delta_max} implies that $\Delta_{ij}^* \leq \Delta_{ij}^\text{max}$ and similarly from \eqref{ineq:relation_delta*_delta_min}, we have $\Delta_{ij}^* \geq - \Delta_{ij}^\text{min}$.
    Therefore, we have $- \Delta_{ij}^\text{min} \leq \Delta_{ij}^* \leq \Delta_{ij}^\text{max}$. Since $\Delta_{ij}^\text{min} \geq 0$, it concludes the proof.
\end{proof}

\subsection{Regret Notions}
To characterize the performance of our algorithms, we consider several regret notions. Let begin with the following formulations, which are the external regret of the max and min players, respectively:
\begin{align*}
    & R_T^{\text{max}} = \max_{i \in S_x} \mathbb{E} \left[\sum_{t=1}^T  A(i,j_t) - A(i_t,j_t)  \right] \\
    & R_T^{\text{min}} = \max_{j \in S_y} \mathbb{E} \left[ \sum_{t=1}^T A(i_t,j_t) - A(i_t, j) \right]
\end{align*}
where $i_t$ and $j_t$ are the actions selected by the row and column player in round $t$, respectively.
$R_T^{\text{max}}$ represents the expected loss due to not selecting the best possible action of the row player against column player's action $j$ while $R_T^{\text{min}}$ captures the expected regret of not choosing the column player's best action for a given action $i$. 

We can now rewrite the regret definitions in terms of the suboptimality gaps:
\begin{align*}
   & R_T^{\text{max}} = \sum_{(i,j) \in S_A} \Delta_{ij}^\text{max}  \mathbb{E}[n_{ij,T}] \\
    & R_T^{\text{min}} = \sum_{(i,j) \in S_A} \Delta_{ij}^\text{min}  \mathbb{E}[n_{ij,T}]
\end{align*}
where $\mathbb{E}[n_{ij,T}]$ denotes the expected number of times the action pair $(i,j)$ is played over $T$ rounds.

In this paper, we analyze two regret notions. The external regret and the Nash regret. 
\begin{definition}[External regret]
\label{eq:externalregret}
The (combined) external regret, which quantifies how much worse the players have performed compared to their best actions in hindsight, is given by the following:
\begin{align}
    R_T = R_T^{\text{max}} + R_T^{\text{min}} =  \sum_{(i,j) \in S_A} \Delta_{ij}  \mathbb{E}[n_{ij,T}] \label{eq:regret-R_T}
\end{align}

\end{definition}

An alternative regret notion is Nash regret, which measures the value of the deviation from the Nash equilibrium:
\begin{definition}[Nash regret]
\label{eq:nashregret}
The Nash regret is expressed as
\begin{align}
    R_T^* = \sum_{(i,j) \in S_A} \Delta_{ij}^* \mathbb{E}[n_{ij,T}] \label{eq:regret-R_T*}.
\end{align}
\end{definition}
The Nash regret, as defined here, can be negative, since the minimizing player might play suboptimally. However, if we are able to bound the Nash regret for the maximizing player, then by symmetry, we can also bound it for the minimizing player, thus effectively bounding its absolute value.

\section{\ETC{} Algorithm}
\label{section:ETC_and_regret}
In this section, we extend the ETC algorithm to a two-person zero-sum game setting, called \ETC{}. Each player selects action from their corresponding finite action sets, which are known. The game is repeated over a given time horizon $T$, and the true payoff matrix $A$ is unknown to both players. Instead, they receive observations of the payoffs based on their chosen actions, which are iid subgaussian random variables. 

Let $k$ represent the number of times to explore each action pair $(i, j) \in S_A$. During the exploration phase, each player samples actions to estimate the expected payoffs, which enables us to construct an empirical payoff matrix using the observed payoffs. Then, in the commit phase, they play according to the pure NE strategy derived from the estimated payoff matrix. Finally, we analyze the expected regret of the algorithm, measuring the performance due to limited information and suboptimal action pairs during the exploration phase.

As a result, the unknown payoff matrix is approximated by the ETC approach, as in \cite{Lattimore_Szepesvari_2020} for a bandit setting. The ETC algorithm consists of two phases: an exploration phase, in which the player randomly samples actions to gather information with a given exploration time $k$, and a commit phase, where the player selects the empirically best action based on the observations. Algorithm \ref{alg:etc_game} presents the implementation of \ETC{}, where each player only observes the payoffs of the action pair they choose during the exploration period, then in the commit phase, the algorithm selects an optimal action pair which refers to the pure NE.

\begin{theorem}
For $1 \leq Nk \leq T$, the Nash regret of Algorithm \ref{alg:etc_game}, interacting with $\sigma$-subgaussian payoffs, is upper bounded by
\begin{align}
    R_T^* \leq k \sum_{i,j} \Delta_{ij}^* + (T - Nk) \sum_{ i,j} \Delta_{ij}^*  \exp{\left( - \frac{ k\Delta_{ij}^2}{ 16 \sigma^2} \right) }
\label{eq:regretbound1}
\end{align}
where $k$ is the exploration time per action pair $(i,j) \in S_A$ and $N$ is the total number of action pairs. 
\label{thm:regretbound_etc_zsg}
\end{theorem}

\begin{proof}[Proof Sketch]
    The total regret consists of two components: the regret from exploration and the regret from exploitation. During the exploration phase, each action pair is played $k$ times. In the exploitation phase, the algorithm plays according to the estimated pure equilibria. In other words, in the exploitation phase, we analyze the loss incurred from playing with a misidentified NE. Thus, the expected number of plays for any suboptimal action pair $(i,j)$ can be written by $\mathbb{E}[n_{ij,T}] = k + (T - Nk) \ \mathbb{P}((i,j) \text{ is identified as NE})$. Once this probability is bounded, the proof is complete. (See Appendix \ref{section:appendix_proof_thm4_1}).
\end{proof}

\begin{algorithm}[ht!]
\caption{\ETC{}}
\label{alg:etc_game}
\begin{algorithmic}[1]
\State \textbf{Input:} $S_A, m, l, k, \sigma, T$
\State \textbf{Initialize:} $\hat{A} = [0]^{m \times l }$
\For{$t = 1$ to $mlk$} \Comment{Exploration Phase}
    \State Explore each $(i,j)$ in $S_A$ $k$ times
    \State Update $\hat{A}$ using \eqref{eq:averagepayoff}
\EndFor
\For{$t = mlk+1$ to $T$}  \Comment{Commit Phase}
    \State Play an equilibrium $(i^*,j^*)$ satisfying:
    \begin{align} 
    \hat{A}(i,j^*) \leq \hat{A}(i^*,j^*) \leq \hat{A}(i^*,j), \quad \forall i \in S_x, j \in S_y
    \notag
    \end{align}
\EndFor
\end{algorithmic}
\end{algorithm}

As mentioned above, the regret comprises two main components, the exploration phase and the loss due to misidentifying the NE. Thus, it is crucial to choose an appropriate exploration time $k$ that balances sufficient exploration and  efficient exploitation. Choosing $k$ too large leads to unnecessary exploration, while setting it too small increases the risk of making suboptimal decisions. Hence, careful tuning of $k$ is essential to minimize overall regret.

Let assume there are two action pairs $(i_1,j)$ and $(i_2,j)$ in the game such that the row player has two actions and the column player has one action. In addition, let suppose the NE is at $(i_1,j)$. Then, we have $\Delta = A(i_1,j) - A(i_2,j) $ and we can write the regret simply as
%R_T^* & \leq k \Delta + (T - 2k) \Delta \exp{\left( - \frac{ k\Delta^2}{ 16 \sigma^2} \right) } \\ 
\begin{align}
    R_T^* \leq k \Delta + T \Delta \exp{\left( - \frac{ k\Delta^2}{ 16 \sigma^2} \right) }.
    \label{ineq:regret_for_two_actionpairs}
\end{align}
Taking the first derivative with respect to $k$ and solving it, we get the following for the exploration time:
\begin{align}
    k = \max \left \{ 1, \left \lceil \frac{16 \sigma^2}{\Delta^2} \ln{\frac{\Delta^2 T}{16\sigma^2}} \right \rceil \right \}
    \label{eq:explorationtimeestimate}
\end{align}
If $\Delta$ is known, we can easily find the necessary exploration time. Therefore, when we put it into the regret bound, it becomes
\begin{align}
    R_T^* \leq \min \left \{ T \Delta, \Delta +  \frac{16 \sigma^2}{\Delta} \left( 1 + \max \left \{ 0, \ln{\frac{\Delta^2 T}{16 \sigma^2}} \right \} \right)
    \right \}
    \label{eq:regretboundmin}
\end{align}
where the first term $T \Delta$ is the worst-case regret and for the other, we combine the regret we obtain in \eqref{ineq:regret_for_two_actionpairs} with the value of the exploration time in \eqref{eq:explorationtimeestimate}. Then, focusing on $\max \left \{ 0, \ln{\frac{\Delta^2 T}{16 \sigma^2}} \right \}$, we obtain $\Delta \geq \frac{4 \sigma}{\sqrt{T}}$. Thus, when we put it in the regret bound \eqref{eq:regretboundmin}, we have the following:
\begin{align}
    R_T^* \leq \Delta + c \sqrt{T}
\end{align}
where $c > 0$ is some constant. If $\Delta \leq 1$, it becomes $R_T^* \leq 1 + c \sqrt{T}$, which refers to an instance-independent bound since it depends only on the time horizon $T$ not on the game instance. Our analysis shows that we obtain a regret bound comparable to the standard bandit setting in \cite{Lattimore_Szepesvari_2020}, indicating that the ETC-based algorithm performs effectively even in the zero-sum game setting.

\section{\ETCAE{} Algorithm}
\label{section:elimination_and_regret}
In this section, we extend Algorithm \ref{alg:etc_game} by incorporating adaptive elimination, called \ETCAE{}. We use the concept of $\varepsilon$-Nash Equilibrium ($\varepsilon$-NE), which approximately satisfies the standard NE condition, to eliminate action pairs. Specifically, if a pure NE exists at $(i^*,j^*)$, then it must satisfy the condition given in \eqref{eq:nasheqproperty} while $\varepsilon$-NE satisfies the NE conditions within a tolerance level $\varepsilon$, making it a suitable criterion in learning settings where exact equilibrium identification is challenging. 

Based on this criterion, we introduce an elimination method in Algorithm \ref{alg:etc_with_action_elimination}, inspired by \cite{auer2010ucb}, to reduce unnecessary exploration of clearly suboptimal action pairs.  Specifically, action pairs that do not satisfy the $\varepsilon$-NE property are eliminated from further play, which enables a more efficient learning process by concentrating exploration on approximately optimal action pairs.

\begin{definition}
    The action pair $(i^*,j^*)$ is an $\varepsilon$-Nash Equilibrium ($\varepsilon$-NE) if the following condition holds for any $\varepsilon > 0$ and for all $i \in S_x$ and $j \in S_y$:
    \begin{align}
     A(i,j^*) - \varepsilon \leq A(i^*,j^*) \leq A(i^*,j) + \varepsilon.
     \label{eq:epsilon-NE-property}
    \end{align}
\end{definition}

For the exploration time of the algorithm, let just go back to the regret bound in \eqref{eq:regretbound1} and assume that there exist some $\Delta^*$ and $\Delta$ such that $\Delta^* \geq \Delta_{ij}^*$ and $\Delta \leq \Delta_{ij}$ for all $(i,j) \in S_A$. Then, when we put them into the regret bound and take its derivative with respect to $k$, we obtain a result similar to that of \eqref{eq:explorationtimeestimate}.

Since the true suboptimality gaps $\Delta$ are unknown, the algorithm adopts a decreasing $\Delta$ approach over rounds. It is motivated by the intuition that the remaining action pairs become closer to the true NE as suboptimal action pairs are progressively eliminated. Consequently, the suboptimality gaps among the remaining action pairs are expected to decrease over time. In other words, as the rounds progress, we expect that the players get closer to true NE, which means that they tend to repeat better action pairs more often. We can afford to use smaller $\Delta$ values to explore action pairs near the NE more thoroughly, since a  smaller $\Delta$ leads to a longer exploration time $k$.

The algorithm schedules $\Delta_{ij}$ across rounds using $\hat{\Delta}_t$, which guides both the exploration time and the tolerance level for the $\varepsilon$-NE condition.  When $\hat{\Delta}_t$ is large, the action pairs are played fewer times and more action pairs are kept during elimination. As $\hat{\Delta}_t$ decreases, the algorithm focuses exploration on pairs closer to equilibrium and eliminates clearly suboptimal ones. Thus, as the exploration time, we can clearly use the following:
\begin{align}
    k_t = \left \lceil \frac{16 \sigma^2}{\hat{\Delta}_t^2} \ln{ \left( \frac{\hat{\Delta}_t^2 T}{16 \sigma^2} \right)} \right \rceil
\label{eq:explorationtime_k}
\end{align} 
In order to perform updates over rounds, an initial estimate is required. When payoffs are bounded within a known interval, such as $[0,1]$, it is common practice to initialize the estimated suboptimality gap $\hat{\Delta}_t$ with value 1, representing the maximum possible difference between arm rewards as in \cite{auer2010ucb}. However, in our setting, the payoffs are assumed to be $\sigma$-subgaussian, which implies that there is no strict upper bound on the suboptimality gaps, we only have $\Delta_{ij} \geq 0$. Since no empirical estimates are available at the initialization step, it is not feasible to use a bound for the expected value of the maximum of $\sigma$-subgaussian random variables to guide the choice of $\hat{\Delta}_t$. To address this, we initialize $\hat{\Delta}_t$ with $4\sigma$, which provides a practical starting point for the analysis.

We set $\hat{\Delta}_t = 2^{-t+2} \sigma$ in each round $t$, where $t$ ranges from $0$ to $\left\lfloor \frac{1}{2} \log_2 \frac{T}{ e} \right\rfloor$, which allows a nonnegative exploration time $k_t$ and tolerance level $\varepsilon_t$. We have $\hat{\Delta}_t \approx \sqrt{ (16 \sigma^2/ k_t) \ln{ (\hat{\Delta}_t^2 T/ 16 \sigma^2 )}}$ from \eqref{eq:explorationtime_k}, which implies $\ln{ (\hat{\Delta}_t^2 T / 16 \sigma^2 )} \geq 0$. It gives us $\hat{\Delta}_t \geq \sqrt{16 \sigma^2 / T}$ and our initialization of $\hat{\Delta}_t$ remains consistent with this observation. Then, using $2^{-t+2} \sigma \geq \sqrt{16 \sigma^2 / T} $ and after some calculations, we obtain $t \leq \frac{1}{2} \log_2 T$. The division by $e$ is a technical adjustment, which makes the bound safer in the analysis; otherwise, the number of plays would be zero in the last round. 

For the property of $\varepsilon$-NE, we define the tolerance parameter as follows:
\begin{align}
    \varepsilon_t = \sqrt{\frac{ 4 \sigma^2 }{k_t} \ln{\left(\frac{\hat{\Delta}_t^2 T}{16 \sigma^2}\right) }}
\label{eq:epsilont}
\end{align}
where $\varepsilon_t$ reflects how much players are allowed to deviate from the equilibrium and the tolerance over time gradually decreases to ensure convergence towards the NE. Substituting condition $k_t\geq \frac{16 \sigma^2}{\hat{\Delta}_t^2} \ln{ \left( \frac{\hat{\Delta}_t^2 T}{16 \sigma^2} \right)}$ into the equation, where $k_t$ is defined in \eqref{eq:explorationtime_k}, we obtain $\varepsilon_t \leq \hat{\Delta}_t / 2$. This bound indicates that as the number of rounds $t$ increases, the tolerance $\varepsilon_t$ becomes smaller, forcing the system to narrow the set of action pairs by progressively eliminating those that do not approximate a NE. As a result, over time, the algorithm keeps fewer action pairs, effectively focusing the exploration on action pairs that are close to the optimal one.

Since we assume that a unique pure NE exists in the game, a well estimated game matrix $\hat{A}$ should also exhibit the properties necessary to support the equilibrium. Specifically, $\hat{A}$ must satisfy the conditions required for the $\varepsilon$-NE in \eqref{eq:epsilon-NE-property} to hold. Based on this principle, Algorithm \ref{alg:etc_with_action_elimination} systematically eliminates action pairs that fail to satisfy this condition, as such pairs cannot be near to the equilibrium. By reducing the set of action pairs to include only the pairs that meet this criterion, it ensures that the remaining action pairs are the only ones that could potentially be near the NE; thus, it enables to focus on the most relevant action pairs. Consequently, we expect that the algorithm converges to an accurate approximation of the NE.

\begin{theorem}
\label{thm:regretbound_actionpairelimination}
For $1 \leq ml \ll T$, the Nash regret of Algorithm \ref{alg:etc_with_action_elimination}, interacting with $\sigma$-subgaussian payoffs, is upper bounded by
\begin{align}
    R_T^*  \leq & \sum_{(i,j) \in S_{A_1}} \Delta_{ij}^*  \left( 1 + \frac{768 \sigma^2}{\Delta_{ij}^2} + \frac{256 \sigma^2} {\Delta_{ij}^2} \ln{\left(\frac{\Delta_{ij}^2 T}{256 \sigma^2}\right)} \right) + \sum_{(i,j) \in S_{A_2}} \left( \Delta_{ij}^* \frac{512 \sigma^2}{\lambda^2} \right) +  \max_{(i,j) \in S_{A_2} } \Delta_{ij}^* T
\label{regret_bound_action_elimination_alg}
\end{align}
where $\lambda \geq 4 \sigma \sqrt{e/T}$, and $S_{A_1} = \{ (i,j) \in S_A: \Delta_{ij} > \lambda \}$ and $S_{A_2} = \{ (i,j) \in S_A: 0 < \Delta_{ij} \leq \lambda \}$ are two subsets of the action pairs.
\end{theorem}

\begin{proof}[Proof Sketch]
    We first define $t_{ij} = \min \{ t: \hat{\Delta}_t < \Delta_{ij} / 2 \}$ as the earliest round satisfying $\hat{\Delta}_t < \Delta_{ij} / 2$. Thus, we consider four distinct cases. First, we analyze the suboptimal action pairs which are not eliminated in round $t_{ij}$ or before. Next, we examine the suboptimal pairs that are eliminated in the round $t_{ij}$ or earlier, allowing us to bound the number of plays with suboptimal action pairs. The third case examines the scenario where the optimal action pair is eliminated in $t_{ij}$. Finally, we consider the case in which a suboptimal action pair remains as a unique one in the game. (See Appendix \ref{section:appendix_proof_thm5_1})
\end{proof}

Moreover, if we consider another regret approach in \eqref{eq:regret-R_T}, we obtain the following regret bound. 

\begin{theorem}
\label{thm:regretbound2_actionpairelimination}
The external regret incurred by Algorithm \ref{alg:etc_with_action_elimination} when interacting with $\sigma$-subgaussian payoffs is bounded as following:
\begin{align}
    R_T \leq & \sum_{(i,j) \in S_{A_1}} \left( \Delta_{ij} +  \frac{768 \sigma^2}{\Delta_{ij}} + \frac{256 \sigma^2} {\Delta_{ij}} \ln{\left(\frac{\Delta_{ij}^2 T}{256 \sigma^2}\right)} \right) + \sum_{(i,j) \in S_{A_2}} \left( \frac{512 \sigma^2}{\lambda} \right) +  \lambda T .
\label{regret_bound2_action_elimination_alg}
\end{align}
\end{theorem}

\begin{proof}
Although the regret notion differs, this does not affect the underlying regret analysis. Thus, the previous theoretical bounds can be extended using Lemma \ref{lemma:relationbetw_delta*_and_delta}, which completes the proof.
\end{proof}

\begin{algorithm}[ht!]
\caption{\ETCAE{}}
\label{alg:etc_with_action_elimination}
\begin{algorithmic}[1]
\State \textbf{Input:} $S_A, m, l, \sigma, T$
\State \textbf{Initialize:} $\hat{\Delta}_0 = 4 \sigma$, $S_0 = S_A$ and $\hat{A} = [0]^{m \times l }$
\For{$t=0,1,2,...,\left\lfloor \frac{1}{2} \log_2 \frac{T}{e} \right\rfloor$} 
    %\State \textit{Action Pair Selection:}
    \If{$\mathbf{|S_t|>1}:$} %\Comment{Exploration Phase}
        \State Explore each $(i,j)$ in $S_t$ $k_t$ times in \eqref{eq:explorationtime_k}
        \State Update $\hat{A}$ using  \eqref{eq:averagepayoff}
    \Else  %\Comment{Commit Phase}
        \State Play with a unique action pair in $S_t$ until step $T$ 
    \EndIf
    %\State \textit{Action Pair Elimination:}
    \State Remove all $(i,j)$ from $S_t$ \textbf{NOT} satisfying: %\Comment{Elimination Phase}
    \begin{align}
        \hat{A}(i',j) - \varepsilon_t \leq \hat{A}(i,j) \leq \hat{A}(i,j') + \varepsilon_t, \forall i', \forall j' \text{ for $\varepsilon_t$ defined in \eqref{eq:epsilont}}
        \notag
    \end{align}
    \State $S_{t+1} \leftarrow S_t$
    \State $ \hat{\Delta}_{t+1} \leftarrow \frac{\hat{\Delta}_{t}}{2}$ where $\hat{\Delta}_t = 2^{-t+2} \sigma$
\EndFor
\end{algorithmic}
\end{algorithm}

We simplify the regret bound to $O\left(\frac{\log (T \Delta^2)}{\Delta}\right)$, as the logarithmic term dominates since by setting $\lambda = \sqrt{16\sigma^2 e / T}$, we ensure that the term $\lambda T$ is bounded by $\sqrt{16 \sigma^2 e T} $, which is at most $\frac{e}{\Delta_{ij}}$ when $\Delta_{ij} \leq \lambda$. Our results are consistent with the regret bound derived by \cite{auer2010ucb}. 

Compared to our result, \cite{ito2025instance} achieves a regret bound of $O \left( \left(\sum_{i \neq i^*} \frac{1}{\Delta_i} + \sum_{j \neq j^*} \frac{1}{\Delta'_j}\right) \log T \right)$ where $\Delta_i = A(i^*,j^*) - A(i,j^*)$ and $\Delta'_j = A(i^*,j) - A(i^*,j^*)$ denote the gaps between pure NE and the payoff obtained by deviating while the opponent plays its equilibrium action. Although the bound is not directly comparable to ours, as the settings and the definition of suboptimality gaps differ, it currently presents the only instance-dependent regret analysis for general $m \times l$ zero-sum matrix games. Our regret bound is tighter and our analysis relies on a more refined notion of the suboptimality gap $\Delta$, as it exhibits a sharper dependence on the gap parameter and avoids the cumulative sum of suboptimality gaps that appears in \cite{ito2025instance}. Furthermore, our elimination algorithm specifically incorporates the extended pure NE property through $\varepsilon$ in a centralized setting, which explains why we achieve a tighter regret bound for a TPZSG setting where both players have a pure strategy; this is further supported by the simulation results in Section \ref{section:experiments}.

As a result, we have $R_T^* \leq R_T$, where $R_T^*$ evaluates its performance against the NE, capturing the cumulative loss relative to the equilibrium strategy, while the regret analysis of $R_T$ allows us to consider the losses of both players with respect to their best individual  responses in each round, providing a broader and more flexible notion of regret. This outcome aligns with the interpretation that the expected regret against the NE establishes a more specific comparison, and thus, it naturally provides a tighter upper bound.

\section{\ETCNUE{} Algorithm}
\label{section:nonuniform_exploration}
In this section, we extend the adaptive elimination algorithm by introducing a variant, \ETCNUE{}. It uses a non-uniform exploration method, which is based on players' strategies, which indirectly reflect the probability of being NE. The key difference from the previous algorithm is that Algorithm \ref{alg:etc_with_nonuniform_exploration} explores action pairs in proportion to the strategies of the players derived from the estimated payoff matrix. Thus, it is employed to reduce the number of plays for suboptimal action pairs more effectively.

In round $t$, we denote the estimated optimal strategy of the row player by $\hat{p}_t$ and the estimated optimal strategy of the column player by $\hat{q}_t$, where they are calculated trough the estimated game matrix $\hat{A}$. We expect these estimated strategies to converge to the pure strategies over time.

The estimated strategies of the players serve as a basis to determine the exploration time in Algorithm \ref{alg:etc_with_nonuniform_exploration}. The joint probability of selecting an action pair $(i,j)$ is estimated by $\hat{P}_{ij,t} = \hat{p}_{i,t} \hat{q}_{j,t}$, which satisfies $\sum_{(i,j) \in S_A} \hat{P}_{ij,t} = 1$, where $(\hat{p}_t,  \hat{q}_t)$ is a mixed NE in round $t$. Consequently, in Algorithm \ref{alg:etc_with_nonuniform_exploration}, the exploration time for each action pair $(i,j)$ is proportional to $\hat{P}_{ij,t}$, allowing action pairs with higher estimated probabilities to be played more frequently. 

We incorporate the strategies of the players $\hat{p}_t$ and $\hat{q}_t$ together with $k_t$ in \eqref{eq:explorationtime_k}. Specifically, in each round $t$, this involves estimating the optimal strategies for each player by solving a linear system using the estimated payoff matrix $\hat{A}$. Thus, the exploration time is given by the following: 
\begin{align}
    k_{ij,t} = k_t + \left \lceil \frac{16 \sigma^2 \hat{P}_{ij,t}}{\hat{\Delta}_t^2} \ln{(\frac{\hat{\Delta}_t^2 T}{16 \sigma^2})} \right \rceil  
\label{eq:nonuniform_exploration_time}
\end{align}
Thus, the algorithm explores each action pair at least $k_t$ times. Since players’ strategies are expected to converge to pure ones over time, the Algorithm \ref{alg:etc_with_nonuniform_exploration} progressively focuses on playing the optimal action pair more. This ensures that action pairs with higher estimated probabilities are explored more frequently, while those with lower probabilities are selected less often. Thus, these algorithms achieve a more efficient balance between exploration and exploitation by reducing unnecessary plays on suboptimal action pairs, which accelerates convergence toward optimal action pair. 

We note that we set $t = \left\lfloor \frac{1}{4} \log_2 \frac{T}{e} \right\rfloor$ as the last round because unlike the Algorithm \ref{alg:etc_with_action_elimination}, which assigns an exploration time of $k_t$, the current algorithms allow exploration up to $2 k_t$ times, ensuring that each action pair is explored at least $k_t$ times. As a result, due to this extended exploration phase, the number of rounds is reduced by half.

\begin{algorithm}[ht]
\caption{\ETCNUE{}}
\label{alg:etc_with_nonuniform_exploration}
\begin{algorithmic}[1]
\State \textbf{Input:} $S_A, m, l, \sigma, T $ 
\State \textbf{Initialize:} $\hat{\Delta}_0 = 4 \sigma$, $S_0 = S_A$ and $\hat{A} = [0]^{m \times l }$
\For{$t=0,1,2,...,\left\lfloor \frac{1}{4} \log_2 \frac{T}{e} \right\rfloor$} 
    \State Run Algorithm \ref{alg:etc_with_action_elimination} with exploration time $k_{ij,t}$ in \eqref{eq:nonuniform_exploration_time}
    \State Calculate optimal strategies $\hat{p}_t$ and $\hat{q}_t$ based on $\hat{A}$
    \State Update $\hat{P}_{ij,t+1}$
\EndFor
\end{algorithmic}
\end{algorithm}

\begin{theorem}
\label{thm:regretbound_nonuniformexploration}
For $1 \leq ml \ll T$, the Nash regret of Algorithm \ref{alg:etc_with_nonuniform_exploration} is upper bounded by
\begin{align}
    R_T^*  \leq & \sum_{(i,j) \in S_{A_1}} \Delta_{ij}^*  \left( 1 + \frac{768 \sigma^2}{\Delta_{ij}^2} + \frac{512 \sigma^2} {\Delta_{ij}^2} \ln{\left(\frac{\Delta_{ij}^2 T}{256 \sigma^2}\right)} \right) + \sum_{(i,j) \in S_{A_2}} \left( \Delta_{ij}^* \frac{512 \sigma^2}{\lambda^2} \right) + \max_{(i,j) \in S_{A_2} } \Delta_{ij}^* T
\label{regret_bound_nonuniform_exploration_alg}
\end{align}
where $\lambda \geq 4 \sigma \sqrt[4]{e/T}$ and, $S_{A_1} = \{ (i,j) \in S_A: \Delta_{ij} > \lambda \}$ and $S_{A_2} = \{ (i,j) \in S_A: 0 < \Delta_{ij} \leq \lambda \}$  are two subsets of the action pairs.

\end{theorem}

\begin{proof}[Proof Sketch]
    Having the same structure as Algorithm \ref{alg:etc_with_action_elimination} and ensuring that each action pair is explored at least $k_t$ times enables us to use the same proof steps; however, this modification affects the bound on the number of plays when the algorithm eliminates a suboptimal action pair in round $t_{ij}$. Detailed explanation is given in Appendix \ref{section:proof_of_thm_nonuniform_exp}. We note that since we have larger $\lambda$ here, it makes this bound tighter compared to the results of Algorithm \ref{alg:etc_with_action_elimination}.
\end{proof}

In conclusion, Algorithm \ref{alg:etc_with_nonuniform_exploration} aims that the strategies of players converge to the corresponding pure strategies over time, which allows more focused exploration on actions near the pure NE and thus enables to eliminate suboptimal action pairs more effectively. In addition, the regret analysis of Algorithm \ref{alg:etc_with_nonuniform_exploration} provides a tighter upper bound, highlighting the improved efficiency of the algorithm in learning pure NE.

\section{Experiments}
\label{section:experiments}
In this section, we present a set of simulations to support our theoretical findings and demonstrate the performance of the proposed algorithms. The experiments are conducted with a fixed time horizon $T=10^3$ for Figure \ref{fig:etc_vs_upperbound} and $T=10^4$ for other simulations. Additional details are provided in Appendix \ref{section:appendix_experiments}.

\begin{figure}[ht]
        \centering
        \includegraphics[width=0.45\linewidth]{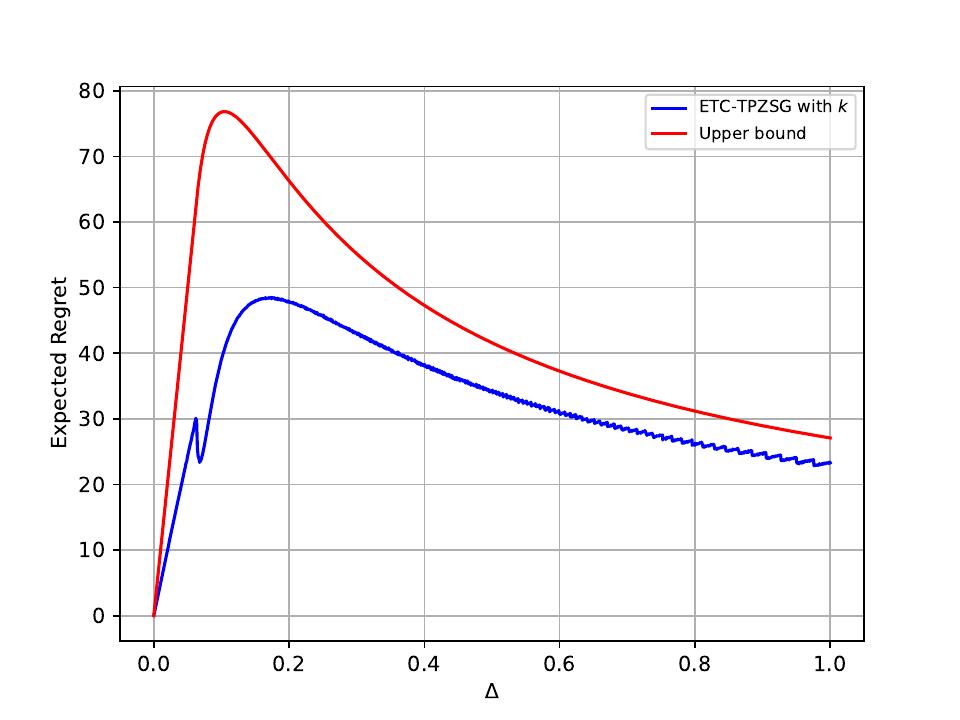}
        \caption{The expected regret of \ETC{} with $k$ in \eqref{eq:explorationtimeestimate} and the upper bound in \eqref{eq:regretboundmin}}
        \label{fig:etc_vs_upperbound}
\end{figure}

\begin{figure*}[ht]
    \centering
    \begin{minipage}[t]{0.45\textwidth}
        \centering
        \includegraphics[width=1\linewidth]{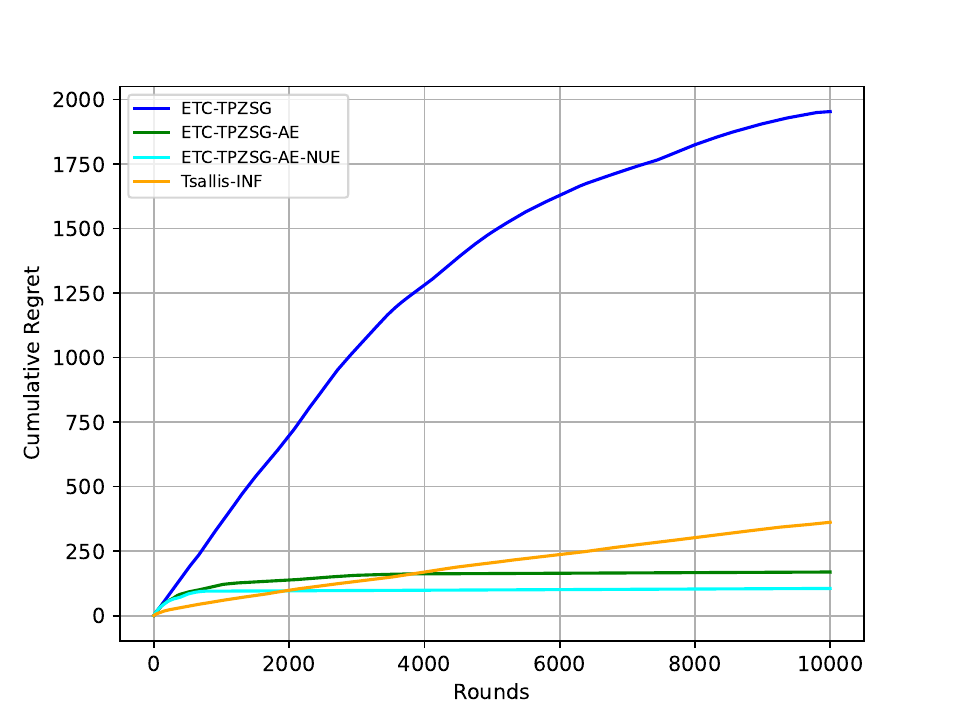}
        \caption{The cumulative regrets of proposed algorithms and Tsallis-INF with large $\Delta$}
        \label{fig:cum_regrets_of_algorithms_large_delta}
    \end{minipage}
    \hfill
    \begin{minipage}[t]{0.45\textwidth}
        \centering
        \includegraphics[width=1\linewidth]{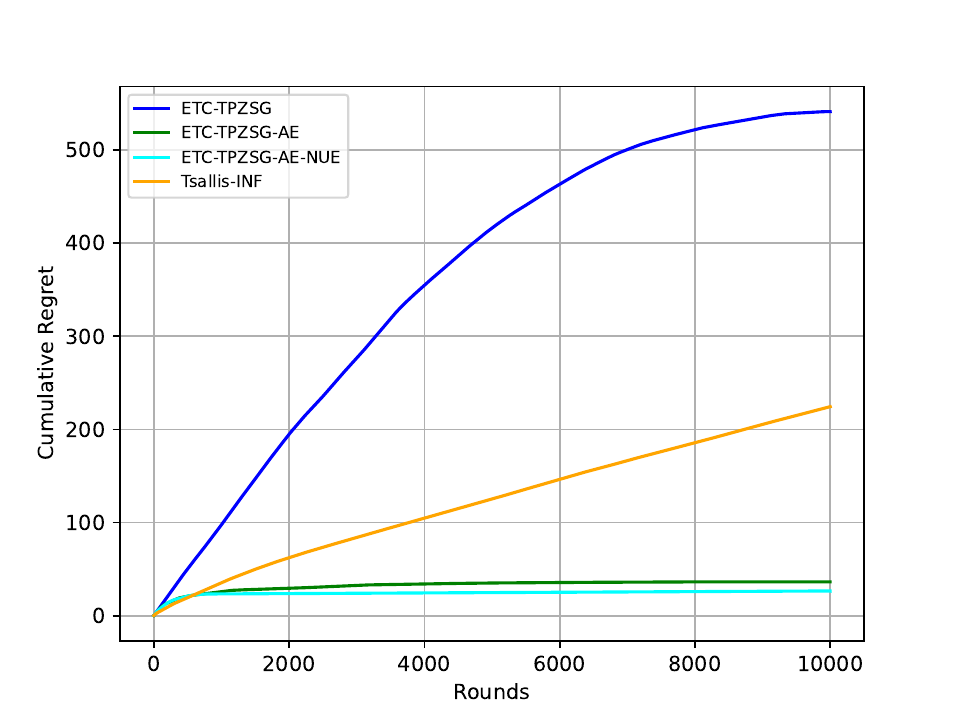}
        \caption{The cumulative regrets of proposed algorithms and Tsallis-INF with small $\Delta$}
        \label{fig:cum_regrets_of_algorithms_small_delta}
    \end{minipage}
\end{figure*}

Figure \ref{fig:etc_vs_upperbound} shows the expected regret, averaged over $10^5$ runs, of \ETC{} using $k$ in \eqref{eq:explorationtimeestimate} and the upper bound in \eqref{eq:regretboundmin}.  We consider a setting with $N=2$, where the row player has two actions, the first corresponds to NE with mean zero and the second with $-\Delta$, and the column player has one action. The suboptimality gap $\Delta$ varies from 0 to 1 and $\sigma=0.5$. The fluctuation around $\Delta = 0.1$ arises since, up to this point, the exploration time $k$ equals one, after which it increases suddenly. The results align with the bound in Theorem \ref{thm:regretbound_etc_zsg}.

In Figures \ref{fig:cum_regrets_of_algorithms_large_delta} and \ref{fig:cum_regrets_of_algorithms_small_delta}, we compare the proposed algorithms and Tsallis-INF \cite{ito2025instance}, very recent method with instance-dependent bounds for TPZSGs, based on cumulative regret, computed as the absolute difference between the game value and the payoff from the action played to avoid negative values, averaged over $10^2$ runs. We use a $2 \times 2$ game matrix by Gaussian payoffs with variance $\sigma^2$ and $k$ exploration rounds per action pair for \ETC{} is randomly selected from a predefined list. We note that a smaller $\sigma$ leads to a smaller $\Delta$. 

We evaluate the algorithms using the payoff matrices with large and small $\Delta$ in Figures  \ref{fig:cum_regrets_of_algorithms_large_delta} and \ref{fig:cum_regrets_of_algorithms_small_delta}, respectively, to examine how $\Delta$ affects the performance of the algorithms. This comparison highlights the importance of instance-dependent regret analysis. The results show that the algorithms incorporating elimination methods achieve lower regret than the \ETC{} and Tsallis-INF algorithms, highlighting the effectiveness of adaptive elimination and exploration. 

On the other hand, when $\Delta$ is small, the Tsallis-INF algorithm fails to correctly identify pure NE as we see an increase in its cumulative regret in Figure \ref{fig:cum_regrets_of_algorithms_small_delta}. In contrast, Algorithms \ref{alg:etc_with_action_elimination} and \ref{alg:etc_with_nonuniform_exploration} perform better.
Thus, we also emphasize that our algorithms preserve the pure NE property, which partly explains why they perform better than Tsallis-INF in games that admit pure NE. Experimental details and further results with different game matrices are provided in Appendix \ref{section:appendix_experiments}.

\section{Discussion}
\label{section:discussion}
We investigate a TPZSG setting with bandit feedback, where the true payoff matrix is unknown and must be learned through player interactions. This setting is challenging, as players must estimate payoffs while making strategic decisions with strictly opposing objectives. We adopt the ETC algorithm due to its simplicity, widespread use and lack of prior analysis in this context. We integrate it to an adaptive elimination method, allowing the systematic removal of suboptimal action pairs, thereby improving convergence to the equilibrium and reducing unnecessary exploration on suboptimal ones. Finally, we combine the adaptive elimination algorithm with a non-uniform exploration approach to explore action pairs more effectively. A key contribution of our work is the derivation of instance-dependent upper bounds on the expected regret for proposed algorithms, which has received limited attention in the literature on zero-sum games.

Although our study focuses on pure strategy learning in a zero-sum game with bandit feedback and provides instance-dependent regret bounds, several directions remain open for future study. An interesting extension is to explore games with a mixed strategy equilibrium, which is more challenging because of the convergence dynamics to true strategies, and also more complex to analyze regret compared to a pure strategy setting. Another direction is to analyze general-sum games, where the player interactions are not strictly adversarial, which requires to consider additional factors to learn optimal strategies.

Finally, an important and relatively unexplored direction is fairness in zero-sum games. For example, introducing mechanisms that ensure similar action pairs are explored equally can promote fairness in strategy estimation. This type of fair play mechanism could be essential in TPZSG settings, thus it can promote balanced exploration and improve overall strategy estimation. On the other hand, by integrating fairness based algorithms or constraints, it may be possible to generate game environments that ensure equal opportunities for all players.

\paragraph{Acknowledgments.} The authors would like to thank Debabrota Basu, Thomas Kleine Buening and Ronald Ortner for fruitful discussions on this topic.

% References
\bibliographystyle{plain} 
\bibliography{references}

\newpage
\appendix

\section{Useful Properties}
In this section, we present several properties that are used throughout the proofs. Moreover, this section collects some supplementary results from \cite{Lattimore_Szepesvari_2020}. These results concern the standard properties of subgaussian random variables. They are included here for completeness and easy reference, as they play a critical role in our regret analyses.
\begin{lemma}
\label{lemma:maxab_greater_sumab}
    For any $a, b > 0$, we can write the following:
    \begin{align}
        \max \{ a^2, b^2 \} \geq \frac{a^2 + b^2}{2}
    \end{align}
\end{lemma}
\begin{proof}
    If we assume that their maximum is $a^2$, then we have $a^2 \geq a^2 + b^2 / 2$, which implies $a^2 \geq b^2$. Similarly, if $\max \{ a^2, b^2 \} = b^2$, it gives us $b^2 \geq a^2$. They are true by assumptions, and thus we conclude the proof.
\end{proof}

\begin{lemma}
\label{lemma:a2+b2_greater_a+b2}
    For any $a, b > 0$, the following holds:
    \begin{align}
        a^2 + b^2 \geq \frac{(a+b)^2}{2}
    \end{align}
\end{lemma}
\begin{proof}
    The proof follows from the fact that $(a-b)^2 \geq 0$.
\end{proof}

\begin{lemma}
\label{lemma:intersectionofevents_less_oneevent}
    For events $e_1, e_2, \cdots, e_n$, we have the following for any $i \in \{ 1, 2, \cdots, n\}$:
    \begin{align}
        \mathbb{P}(e_1 \cap e_2 \cap \cdots \cap e_n) \leq \mathbb{P}(e_i).
    \end{align}
\end{lemma}
\begin{proof}
    For any $i \in \{ 1, 2, \cdots, n\}$, we observe that $e_1 \cap e_2 \cap \cdots \cap e_n \subseteq e_i$. Thus, the proof comes from the fact that an intersection of events cannot be more likely than any of the events involved.
\end{proof}

\begin{lemma}[{\cite[Theorem 5.3]{Lattimore_Szepesvari_2020}}]
\label{lemma:subgaussian_tail_bound}
If $X$ is $\sigma$-subgaussian random variable, for any $\epsilon \geq 0$,
\begin{align}
    \mathbb{P}(X \geq \epsilon) \leq \exp{\left( -\frac{\epsilon^2}{2\sigma^2}\right)}.
    \label{eq:subgaussian}
\end{align}
\end{lemma}

\begin{lemma}[{\cite[Lemma 5.4]{Lattimore_Szepesvari_2020}}]
\label{lemma:subgaussian_properties}
If $X$ is $\sigma$-subgaussian and $X_1$ and $X_2$ are independent with $\sigma_1$ and $\sigma_2$ subgaussian parameter, respectively, then we can write the followings:
\begin{enumerate}[label=(\alph*)]
\item $cX$ is $|c| \sigma$-subgaussian for all $c \in \mathbb{R}.$
\item $X_1 + X_2$ is $\sqrt{\sigma_1^2 + \sigma_2^2}$-subgaussian.
\end{enumerate}
\end{lemma}

\section{Proof of Theorem \ref{thm:regretbound_etc_zsg}}
\label{section:appendix_proof_thm4_1}
To analyze the Nash regret of the Algorithm \ref{alg:etc_game}, we begin by decomposing the total expected regret into two phases: the exploration phase and the commit phase. During exploration, each action pair is played a fixed number of times $k$ to estimate their mean payoffs, potentially incurring regret if suboptimal action pairs are played. Once the algorithm commits to the empirically best action pair, which is the pure NE, the regret accumulates only if this action pair is not the true optimal one. Our goal is to bound the expected regret by quantifying the probability of committing to a suboptimal action pair, which we can call it a misidentified NE, based on the estimates of their payoffs obtained during the exploration phase.

Let $(i',j')$ denote a misidentified NE, so we aim to identify an action pair  $(i',j')$, which appears better than $(i^*,j^*)$ based on the estimated matrix. To analyze the probability of playing with a misidentified NE, we consider whether the following two events, $E_1$ and $E_2$, occur:
\begin{equation*}
E_1: \hat{A}(i,j') \leq \hat{A}(i',j'), \quad E_2: \hat{A}(i',j') \leq \hat{A}(i',j), \quad \forall i \in S_x, \forall j \in S_y
\end{equation*}
and we want to find a bound for $\mathbb{P}(E_1 \cap E_2)$ because these two conditions must be satisfied to consider $(i'.j')$ as a NE. 

Since the events $E_1$ and $E_2$ are not independent, we consider two different approaches to find an upper bound for $\mathbb{P}(E_1 \cap E_2)$. We then select the approach that gives the tighter bound, as it provides a more accurate estimate of this probability. The first approach uses the fact that $\mathbb{P}(E_1 \cap E_2) \leq \mathbb{P}(E_1)$ and $\mathbb{P}(E_1 \cap E_2) \leq \mathbb{P}(E_2)$, which together imply $\mathbb{P}(E_1 \cap E_2) \leq \sqrt{\mathbb{P}(E_1) \mathbb{P}(E_2)}$. Alternatively, we can use the inequality $\mathbb{P}(E_1 \cap E_2) \leq \min \{\mathbb{P}(E_1), \mathbb{P}(E_2)\}$, which may provide a tighter bound depending on the values of $\mathbb{P}(E_1)$ and $\mathbb{P}(E_2)$. The choice between these approaches depends on which enables us the smaller upper bound.

Let us start by considering the first approach. We can write it as follows:
\begin{align}
    \mathbb{P}(E_1 \cap E_2) 
    & = \mathbb{P}(\hat{A}(i,j')  \leq \hat{A}(i',j')  \wedge \hat{A}(i',j')  \leq \hat{A}(i',j), \forall i, j) \\
    & \leq \mathbb{P}(\hat{A}(i_1,j')  \leq \hat{A}(i',j')  \wedge \hat{A}(i',j')  \leq \hat{A}(i',j_1) )  \label{ineq:prob_intersectionofevents_to_one_event} \\
    & \leq \sqrt{ \mathbb{P}(\hat{A}(i_1,j')  \leq \hat{A}(i',j') ) \ \mathbb{P}( \hat{A}(i',j')  \leq \hat{A}(i',j_1) ) }   \\
    & = \sqrt{ \mathbb{P}(\hat{A}(i',j') - \hat{A}(i_1,j')     \geq 0) \ \mathbb{P}( \hat{A}(i',j_1) - \hat{A}(i',j')  \geq 0) }  \\
    % & = \sqrt{ \mathbb{P}(\hat{A}(i,j') + A(i,j') - \hat{A}(i',j') - A(i',j')  \leq A(i,j') - A(i',j')) \ \mathbb{P}( \hat{A}(i',j') + A(i',j') -  \hat{A}(i',j) - A(i',j) \leq A(i',j') - A(i',j)) }  \\
    & \leq \sqrt{ \mathbb{P}(\hat{A}(i',j') - A(i',j') - (\hat{A}(i_1,j') - A(i_1,j') )  \geq \Delta_{i'j'}^\text{max}) } \notag \\ &\quad \times \sqrt{ \mathbb{P}( \hat{A}(i',j_1) - A(i',j_1) - (\hat{A}(i',j') - A(i',j') )  \geq \Delta_{i'j'}^\text{min}) }  \label{ineq:greaterprob_withdeltas} \\
    & \leq \sqrt{ \exp{\left( - \frac{ k (\Delta_{i'j'}^\text{max})^2}{4 \sigma^2} \right)} \exp{ \left( - \frac{k (\Delta_{i'j'}^\text{min})^2}{4 \sigma^2}\right)} }  \label{ineq:sabg_property_used}\\
    & = \sqrt{ \exp{\left( - \frac{ k[(\Delta_{i'j'}^\text{max})^2 + (\Delta_{i'j'}^\text{min})^2]}{4 \sigma^2} \right) } }  \\
    & \leq \sqrt{ \exp{\left( - \frac{ k  (\Delta_{i'j'}^\text{max} + \Delta_{i'j'}^\text{min} )^2 }{8 \sigma^2} \right) } } \label{ineq:to_deltas_to_sum}   \\
    & = \sqrt{ \exp{\left( - \frac{ k(\Delta_{i'j'})^2}{ 8 \sigma^2} \right) } } \\
    & = \exp{\left( - \frac{ k(\Delta_{i'j'})^2}{ 16 \sigma^2} \right) }
\label{eq:probofmisidentifiedNE_1}
\end{align} 
where we assume that $A(i_1, j') = \max_{i \in S_x} A(i,j')$  and $A(i', j_1) = \min_{i \in S_x} A(i',j)$. On the other hand, we note $ \Delta_{i'j'}^\text{max} = \max_i A(i,j') - A(i',j')$, $ \Delta_{i'j'}^\text{min} = A(i',j') - \min_j A(i',j)$ and $\Delta_{i'j'} = \Delta_{i'j'}^\text{max} + \Delta_{i'j'}^\text{min}$. 

To obtain \eqref{ineq:prob_intersectionofevents_to_one_event}, we utilize the fact that the probability of the intersection of multiple events is at most that of any individual event, as stated in \ref{lemma:intersectionofevents_less_oneevent}. For step \eqref{ineq:sabg_property_used}, we apply Lemma \ref{lemma:subgaussian_tail_bound} with $ \sqrt{2\sigma^2 / k}$-subgaussian random variable. At this point, we need to show that the differences $\hat{A}(i',j') - A(i',j') - \hat{A}(i_1,j') +  A(i_1,j')$ and $\hat{A}(i',j_1) - A(i',j_1) - \hat{A}(i',j') + A(i',j')$ are $ \sqrt{2\sigma^2 / k}$-subgaussian. To establish this, we utilize Lemma \ref{lemma:subgaussian_properties}, which provides key properties of subgaussian random variables. Specifically, for any $(i,j) \in S_A$, $\hat{A}(i,j) - A(i,j)$ are $\sqrt{\sigma^2 / k}$-subgaussian since during the exploration phase, each action pair is played $k$ times. Then by Lemma \ref{lemma:subgaussian_properties}, the difference of two independent subgaussian random variables with parameters $\sqrt{\sigma^2 / k}$ is subgaussian with parameter $\sqrt{2\sigma^2 / k}$. To obtain the last inequality \eqref{ineq:to_deltas_to_sum}, we apply Lemma \ref{lemma:a2+b2_greater_a+b2}.

As we mentioned earlier, there is an alternative approach to bound $\mathbb{P}(E_1 \cap E_2)$ to find a bound for the probability of a misidentified NE. It involves an inequality based on the minimum of individual probabilities. Similarly, let follow this:  
\begin{align}
    \mathbb{P}(E_1 \cap E_2) 
    & = \mathbb{P}(\hat{A}(i,j')  \leq \hat{A}(i',j')  \wedge \hat{A}(i',j')  \leq \hat{A}(i',j), \forall i, j) \\
    & \leq \mathbb{P}(\hat{A}(i_1,j')  \leq \hat{A}(i',j')  \wedge \hat{A}(i',j')  \leq \hat{A}(i',j_1) ) \\
    & \leq \min \{ \mathbb{P}(\hat{A}(i',j') - \hat{A}(i_1,j')     \geq 0) \ \mathbb{P}( \hat{A}(i',j_1) - \hat{A}(i',j')  \geq 0) \} \\
    & \leq \min \left \{ \exp{\left( - \frac{ k (\Delta_{i'j'}^\text{max})^2}{4 \sigma^2} \right)} , \exp{ \left( - \frac{k (\Delta_{i'j'}^\text{min})^2}{4 \sigma^2}\right)} \right \} \\
    & = \exp{\left( - \frac{ k \max \{ (\Delta_{i'j'}^\text{max})^2, (\Delta_{i'j'}^\text{min})^2 \} }{4 \sigma^2} \right)}  \\
    & \leq \exp{\left( - \frac{ k[(\Delta_{i'j'}^\text{max})^2 + (\Delta_{i'j'}^\text{min})^2]}{8 \sigma^2} \right) } \label{ineq:max_to_sum_deltas}   \\
    & \leq \exp{\left( - \frac{ k  (\Delta_{i'j'}^\text{max}+ \Delta_{i'j'}^\text{min} )^2 }{16 \sigma^2} \right) }    \\
    & \leq \exp{\left( - \frac{ k \Delta_{i'j'}^2}{ 16 \sigma^2} \right) } 
\label{eq:probofmisidentifiedNE_2}
\end{align} 
where to get the inequality \eqref{ineq:max_to_sum_deltas}, we use Lemma \ref{lemma:maxab_greater_sumab}. 

Thus, both approaches provide the same upper bound for the probability of misidentified NE. In the regret analysis, we will focus on the loss due to playing with suboptimal (or not true NE) action pairs. Thus, after playing $T$ times, $1 \leq Nk \leq T$ where $N = ml$ is the number of action pairs, we can write the Nash regret as
\begin{align} 
    R_T^* = \sum_{(i,j) \in S_A} \Delta_{ij}^* \mathbb{E}[n_{ij,T}] 
\end{align}
where $\mathbb{E}[n_{ij,T}]$ is the expected number of plays for the action pair $(i, j)$. We already know that each action pair $(i, j) \in S_A$ is played at least $k$ times for exploration. After the exploration step, we will consider playing the misidentified NE $(i',j')$ instead of the true one $(i^*,j^*)$ because we expect the players to select the NE action pair as optimal after the exploration phase. Since we already have the probability of playing a misidentified NE from \eqref{eq:probofmisidentifiedNE_1} or \eqref{eq:probofmisidentifiedNE_2}, we can write 
\begin{align}
\mathbb{E}[n_{ij,T}] \leq k + (T - Nk) \exp{\left( - \frac{ k\Delta_{ij}^2}{ 16 \sigma^2} \right) }.
\end{align}
Hence, the proof is concluded.
\bbox

\section{Proof of Theorem \ref{thm:regretbound_actionpairelimination}}
\label{section:appendix_proof_thm5_1}
According to the algorithm, $4 \sigma \sqrt{e/T}$ is a critical value for $\hat{\Delta}_t$ since it takes this value in the round $\frac{1}{2} \log_2 \frac{T}{e} $. In this way, let define a threshold parameter as $\lambda \geq 4 \sigma \sqrt{e/T}$, which gives us two subsets of actions pairs such that $S_{A_1} = \{ (i,j) \in S_A: \Delta_{ij} > \lambda \}$ and $S_{A_2} = \{ (i,j) \in S_A: 0 < \Delta_{ij} \leq \lambda \}$. Since $\Delta_{ij}^* \leq \Delta_{ij}$ from Lemma \ref{lemma:relationbetw_delta*_and_delta}, it ensures that the bounds are preserved and it makes sense to use these action pair sets in the analysis. Thus, we can write the Nash regret as 
\begin{align}
    R_T^* = \sum_{(i,j) \in S_{A_1}} \Delta_{ij}^* \mathbb{E}[n_{ij,T}] + \sum_{(i,j) \in S_{A_2}} \Delta_{ij}^* \mathbb{E}[n_{ij,T}]
\end{align}
where $\Delta_{ij}^* = A(i^*, j^*) - A(i,j).$ and $n_{ij,T}$ is the total number of times to play the action pair $(i,j)$ by time $T$. 

To analyze the Nash regret of Algorithm \ref{alg:etc_with_action_elimination}, it is essential to consider various scenarios that may lead to suboptimal outcomes. Each of these scenarios contributes to the total Nash regret, as they either involve the incorrect elimination of the optimal action pair or the unnecessary selection of suboptimal pairs. The former results in discarding the optimal strategy because of insufficient evidence, while the latter leads to spending more time with action pairs that are unlikely to be part of any near optimal equilibrium. 

The algorithm employs a scheduling mechanism for $\Delta_{ij}$, where $\hat{\Delta}_t$ is halved in each round. This approach is necessary because the true values of $\Delta_{ij}$ are unknown to the players. We use $\hat{\Delta}_t$ to determine both the exploration time and the tolerance level in the $\varepsilon$-NE property. Specifically, when $\hat{\Delta}_t$ is large, the algorithm plays each action pair fewer times and keeps more pairs during the elimination step, since the threshold for elimination is relatively loose. As $\hat{\Delta}_t$ decreases, the algorithm allocates more exploration to action pairs that are closer to being part of a NE, because those that are clearly suboptimal have already been eliminated. This adaptive process enables a more precise identification of near-optimal strategies by the $\varepsilon$-NE condition while minimizing regret.

For each suboptimal action pair $(i,j)$, let $t_{ij} = \min \{ t: \hat{\Delta}_t < \Delta_{ij} / 2 \}$ refer to the earliest round such that $\hat{\Delta}_t < \Delta_{ij} / 2$. Specifically, we are interested in the first round where $\hat{\Delta}_t$ satisfies both $\hat{\Delta}_t < \Delta_{ij}^{\text{max}}$ and $\hat{\Delta}_t < \Delta_{ij}^{\text{min}}$, but combining these conditions results in the simplified condition $\hat{\Delta}_t < \Delta_{ij} / 2$. Using the fact that $\hat{\Delta}_{t+1} = \frac{\hat{\Delta}_{t}}{2}$ and the definition of $t_{ij}$, we can write the following:
\begin{align}
    \frac{1}{\hat{\Delta}_{t_{ij}}} \leq \frac{4}{\Delta_{ij}} < \frac{1}{\hat{\Delta}_{t_{ij}+1}}
\label{ineq:Delta_t}
\end{align}

Furthermore, we have the following:
\begin{align}
     \varepsilon_{t_{ij}} = \sqrt{\frac{4 \sigma^2}{k_{t_{ij}}} \ln{\left( \frac{\hat{\Delta}_{t_{ij}}^2 T}{16 \sigma^2} \right)}} \leq \frac{\hat{\Delta}_{t_{ij}}}{2} = \hat{\Delta}_{{t_{ij}}+1} < \frac{\Delta_{ij}}{4}
\label{eq:epsilont_vs_deltat}
\end{align}
where the exploration time $k_t$ is defined in \eqref{eq:explorationtime_k}, which satisfies $k_{t_{ij}} \geq \frac{16 \sigma^2}{\hat{\Delta}_{t_{ij}}^2} \ln{ \left( \frac{\hat{\Delta}_t^2 T}{16 \sigma^2} \right)}$.

%$\varepsilon_{t_{ij}} = \sqrt{\frac{ 4 \sigma^2 }{k_{t_{ij}}} \ln{\left(\frac{\hat{\Delta}_{t_{ij}}^2 T}{16 \sigma^2} \right) }} \leq \frac{\hat{\Delta}_{t_{ij}}}{2} = \hat{\Delta}_{t_{ij}+1} < \frac{\Delta_{ij}}{4}$. 

For simplicity of notation, we write $\hat{A}$ instead of $\hat{A}_t$ for the estimated payoff matrix in round $t$ throughout this section. Since we already include additional terms such as $\varepsilon_t$ that refers to the tolerance level in round $t$, it is clear enough which round is being referenced.

We now analyze the regret contributions by considering the following cases:

\begin{case}
    A suboptimal action pair $(i,j)$ is not eliminated in round $t_{ij}$ or earlier while the optimal action pair is in the set $S_{t_{ij}}$.
\end{case}
Since neither some suboptimal nor the optimal action pair is eliminated, the algorithm fails to discard suboptimal choices. It might converge to an incorrect solution without eliminating suboptimal actions, but since it still maintains the optimal action pair within the action pair selection set. The regret contribution of this case comes from the fact that the algorithm spends time with a worse choice when a better one is already available. 

Let explain briefly in which situations the algorithm eliminates or keeps an action pair. For action pair $(i,j)$, if the $\varepsilon$-NE property does not hold in round $t = t_{ij}$, which implies that at least one of the inequalities fails, then it will be eliminated in round $t_{ij}$. In other words, if there exists $i' \in S_x$ or  $j' \in S_y$, such that  one of the following is true:
\begin{align}
    I_1: \hat{A}(i',j) - \hat{A}(i,j) > \varepsilon_{t}, \qquad I_2: \hat{A}(i,j) - \hat{A}(i,j') > \varepsilon_{t},
\end{align}
then $(i,j)$ is eliminated. The inequality $I_1$ indicates that there exists an alternative action $i'$ for the row player that offers a higher payoff than action $i$ against column action $j$. Similarly, $I_2$ implies that there exists a better action $j'$ for the column player than the action $j$ when playing against the row action $i$. If $I_1$ or $I_2$ holds, the action pair $(i, j)$ cannot be part of the NE, as at least one player has an incentive to deviate. Hence, $(i,j)$ is not a NE with high probability.

On the other hand, to keep an action pair $(i,j)$ the following events must both hold:
\begin{equation}
E_1: \hat{A}(i',j) - \varepsilon_t \leq \hat{A}(i,j), \forall i' \in S_x,  \quad E_2: \hat{A}(i,j) \leq \hat{A}(i,j') + \varepsilon_t, \forall j' \in S_y
\label{events:tokeepactionpair}
\end{equation}
To calculate the probability of keeping an action pair $(i,j)$, denoted by $\mathbb{P}(E_1 \cap E_2)$, we begin by analyzing the event $E_1$. Specifically, for an action pair $(i,j)$ which is not a NE, i.e. there exists at least one $i'$ such that $A(i', j) > A(i, j)$, the probability of $E_1$ can be expressed as follows:
\begin{align}
    \mathbb{P}(E_1)  & = \mathbb{P}(\hat{A}(i, j) - \hat{A}(i', j) + \varepsilon_t \geq 0, \forall i' \in S_x) \\
    & \leq \mathbb{P}(\hat{A}(i, j) - \hat{A}(i_1, j) + \varepsilon_t \geq 0) \label{ineq:forallprobs_to_singleprob} \\
    & = \mathbb{P}(\hat{A}(i, j) - A(i, j) - (\hat{A}(i_1, j) - A(i_1, j) ) \geq \Delta_{ij}^{\text{max}} -  \varepsilon_t) \\
    & \leq  \exp{\left( - \frac{ k_t (\Delta_{ij}^{\text{max}} - \varepsilon_t)^2 }{4 \sigma^2} \right) } \label{ineq:apply_subg_thm_deltamax}
\end{align}
where we assume $A(i_1, j) = \max_{i \in S_x} A(i,j)$  and $A(i, j_1) = \min_{j \in S_y} A(i,j)$. To derive the inequality \eqref{ineq:forallprobs_to_singleprob}, we use the fact that the probability of the intersection of multiple events is always less than or equal to the probability of any of the individual events. Specifically, since we have $m$ actions in $S_x$, for any collection of events $e_1, e_2, \cdots, e_m$, we have $\mathbb{P}(e_1 \cap e_2 \cap \cdots \cap e_m) \leq \mathbb{P}(e_1)$, as in Lemma \ref{lemma:intersectionofevents_less_oneevent}.

Then, we see that $\hat{A}(i, j) - A(i, j) - (\hat{A}(i_1, j) - A(i_1, j)$ is $\sqrt{2 \sigma^2 / k_t}$-subgaussian using subgaussian properties. Thus, we apply Lemma \ref{lemma:subgaussian_tail_bound} to get inequality \eqref{ineq:apply_subg_thm_deltamax} since $\Delta_{ij}^{\text{max}} - \varepsilon_t \geq 0$ where $t$ is round $t_{ij}$ or later.

Similarly, in order to calculate the probability of the event $E_2$, we can write
\begin{align}
    \mathbb{P}(E_2)  & = \mathbb{P}(\hat{A}(i, j') - \hat{A}(i, j) + \varepsilon_t \geq 0, \forall i' \in S_x) \\
    & \leq \mathbb{P}(\hat{A}(i, j_1) - \hat{A}(i, j) + \varepsilon_t \geq 0) \\
    & = \mathbb{P}(\hat{A}(i, j_1) - A(i, j_1) - (\hat{A}(i, j) - A(i, j) ) \geq \Delta_{ij}^{\text{min}} -  \varepsilon_t) \\
    & \leq  \exp{\left( - \frac{ k_t (\Delta_{ij}^{\text{min}} - \varepsilon_t)^2 }{4 \sigma^2} \right) } 
\end{align}
where $A(i, j_1) = \min_{j \in S_y} A(i,j)$. We note that $\hat{A}(i, j_1) - A(i, j_1) - (\hat{A}(i, j) - A(i, j)$ is $\sqrt{2 \sigma^2 / k_t}$-subgaussian. Then, we apply Lemma \ref{lemma:subgaussian_tail_bound}, as we have $\Delta_{ij}^{\text{min}} - \varepsilon_t \geq 0$ for $t=t_{ij}$ or later rounds from the definition of $t_{ij}$.

Therefore, we write the probability of keeping an action pair $(i, j)$ as
\begin{align}
    \mathbb{P}(E_1 \cap E_2) \leq \min \{ \mathbb{P}(E_1), \mathbb{P}(E_2) \} & \leq \min \left \{ \exp{\left( - \frac{ k_t (\Delta_{ij}^{\text{max}} - \varepsilon_t)^2 }{4 \sigma^2} \right) }, \exp{\left( - \frac{ k_t (\Delta_{ij}^{\text{min}} - \varepsilon_t)^2 }{4 \sigma^2} \right) }  \right \} \\
    & = \exp{\left( - \frac{ k_t \max \{ (\Delta_{ij}^\text{max} - \varepsilon_t)^2, (\Delta_{ij}^\text{min} - \varepsilon_t)^2 \} }{4 \sigma^2} \right)} \\
    & \leq \exp{\left( - \frac{ k_t ( (\Delta_{ij}^\text{max} - \varepsilon_t)^2 + (\Delta_{ij}^\text{min} - \varepsilon_t)^2 ) }{8 \sigma^2} \right)}  \label{ineq:maxtosum}\\
    & \leq \exp{\left( - \frac{ k_t (\Delta_{ij} - 2\varepsilon_t)^2 }{16 \sigma^2} \right)} \label{ineq:deltamaxmintodeltaij}\\
    & < \exp{\left( - \frac{ k_t \varepsilon_t^2 }{4 \sigma^2} \right)} \label{ineq:deltaij_toepsilont} \\ 
    &  = \exp{\left( - \frac{ k_t \frac{4 \sigma^2}{k_t} \ln{\left( \frac{\hat{\Delta}_t^2 T}{16 \sigma^2} \right)}}{4 \sigma^2} \right) } \\
    & = \frac{16 \sigma^2}{\hat{\Delta}_t^2 T}.
\label{eq:prob_of_keep_an_action_pair}
\end{align}
For \eqref{ineq:maxtosum}, we apply Lemma \ref{lemma:maxab_greater_sumab}. Then , to write \eqref{ineq:deltamaxmintodeltaij}, we use Lemma \ref{lemma:a2+b2_greater_a+b2}. Using the definition of $t_{ij}$ and \eqref{eq:epsilont_vs_deltat}, $\varepsilon_{t} < \Delta_{ij} / 4$ for rounds $t \geq t_{ij}$. Then, we derive the inequality \eqref{ineq:deltaij_toepsilont}. It is important to note that this probability bound holds only for rounds $t_{ij}$ or later. However, the probability of keeping a suboptimal action pair $(i,j)$ in round $t_{ij}$ also accounts for previous rounds, as it implies that the algorithm has not eliminated this pair in earlier rounds.

Therefore, the probability that a suboptimal action pair $(i,j)$ is not eliminated in round $t_{ij}$ or before is bounded by $\frac{16 \sigma^2}{\hat{\Delta}_{t_{ij}}^2 T}$. Then, the regret contribution is simply bounded by a worst case which is $T \Delta_{ij}^*$ for any suboptimal action pair $(i,j) \in S_{A_1}$.

Hence, using the probability of keeping a suboptimal action pair and summing up over all action pairs, we can write their regret contribution as
\begin{align}
    \sum_{(i,j) \in S_{A_1}} \Delta_{ij}^* T \frac{16 \sigma^2}{\hat{\Delta}_{t_{ij}}^2 T} \leq \sum_{(i,j) \in S_{A_1}} \Delta_{ij}^*  \frac{256 \sigma^2}{\Delta_{ij}^2}
\label{eq:regret_contribution_case1}
\end{align}
where we apply inequality \eqref{ineq:Delta_t} to bound it.

Here, we note that we do not need to consider the action pairs in $S_{A_2}$ because, by the design of the algorithm, the gap $\hat{\Delta}_t$ is guaranteed to be at least around $\lambda$ in the corresponding rounds. If the optimal action pair remains in the game, then all suboptimal action pairs should have already been eliminated by the last round or earlier. This means that suboptimal choices are progressively discarded over time as the algorithm learns. As a result, after all the rounds have been played, only one action pair should remain since we consider the case the optimal action pair $(i^*, j^*)$ remain in the game in addition to elimination of suboptimal ones. This final pair is expected to correspond to the pure NE, representing the best choice for both players with no incentive to change their actions. 

\begin{case}
    A suboptimal action pair $(i,j)$ is eliminated in round $t_{ij}$ or earlier with the optimal action pair in $S_{t_{ij}}$.
\end{case}
It enables us to bound the number of times it is played. Once a suboptimal action pair is eliminated and the optimal action pair $(i^*,j^*)$ remains in the game, it no longer contributes to the regret in next rounds. This is because regret arises only when a suboptimal action is chosen instead of the optimal one. 

On the other hand, we note that there is no need to consider the action pairs $(i,j) \in S_{A_2}$ because we handle the elimination of suboptimal pairs before the final round. Thus, using \eqref{ineq:Delta_t}, each action pair $(i,j)$ is played at most $k_{t_{ij}}$ times such that 
\begin{align}
    k_{t_{ij}} = \left \lceil \frac{16 \sigma^2}{\hat{\Delta}_{t_{ij}}^2} \ln{(\frac{\hat{\Delta}_{t_{ij}}^2 T}{16 \sigma^2})} \right \rceil 
\leq \left \lceil \frac{256 \sigma^2}{\Delta_{ij}^2} \ln{\left(\frac{\Delta_{ij}^2 T}{256 \sigma^2}\right)} \right \rceil .
\end{align}

Then, the regret contribution is expressed by
\begin{align}
    \sum_{(i,j) \in S_{A_1}} \Delta_{ij}^* \left \lceil \frac{256 \sigma^2}{\Delta_{ij}^2} \ln{\left(\frac{\Delta_{ij}^2 T}{256 \sigma^2}\right)} \right \rceil 
    & \leq \sum_{(i,j) \in S_{A_1}} \Delta_{ij}^* \left( 1 + \frac{256 \sigma^2}{\Delta_{ij}^2} \ln{\left(\frac{\Delta_{ij}^2 T}{256 \sigma^2}\right)} \right) \\
    & = \sum_{(i,j) \in S_{A_1}} \Delta_{ij}^* + \Delta_{ij}^* \frac{256 \sigma^2}{\Delta_{ij}^2} \ln{\left(\frac{\Delta_{ij}^2 T}{256 \sigma^2}\right)} \\
\label{eq:regret_contribution_case2}
\end{align}

\begin{case}
    The optimal action pair $(i^*, j^*)$ is eliminated by some suboptimal one $(i, j)$ in round $t_*$ such that $t_{ij} \geq t_*$.
\end{case}
It leads to a misidentification of the NE, implying that an action pair $(i, j)$ should be better than the optimal one based on the current estimates. By the $\varepsilon$-NE property, we clearly have the following: 
\begin{equation*}
\hat{A}(i^*, j) - \varepsilon_{t_*} \leq \hat{A}(i, j) \leq \hat{A}(i, j^*) + \varepsilon_{t_*}.
\end{equation*} 
However, it is not enough to keep a suboptimal action pair $(i, j)$. If the algorithm keeps a suboptimal action pair $(i,j)$ up to round $t_*$, the events specified in \eqref{events:tokeepactionpair} must be satisfied with the parameter $\varepsilon_{t_*}$. Furthermore, we note that if the optimal action pair $(i^*, j^*)$ is eliminated in round $t_*$, it must fail to satisfy the $\varepsilon$-NE property as the following:
\begin{equation*} 
     \hat{A}(i,j^*) - \hat{A}(i^*,j^*) > \varepsilon_{t_*} \quad \text{or/and} \quad \hat{A}(i^*,j^*) - \hat{A}(i^*,j) > \varepsilon_{t_*}.
\end{equation*} 
That is, under the estimated payoff matrix $\hat{A}$, there exists an action $i$ that is estimated to yield a higher payoff than $i^*$ for the row player, and an action $j$ that is estimated to be more favorable than $j^*$ for the column player.

We observe that all action pairs $(i,j)$ with $t_{ij} < t_*$ have already been eliminated in round $t_{ij}$ or earlier, which we already consider this condition in the previous case. Then, the optimal action pair can only be eliminated by a suboptimal action pair $(i, j)$ in round $t_*$ such that $t_{ij} \geq t_*$ since action pair $(i, j)$ must remain under consideration at the elimination round of the optimal action pair. This condition also implies that $\hat{\Delta}_{t_{ij}} \leq \hat{\Delta}_{t_*}$ because of the decreasing behavior of $\hat{\Delta}_t$ across rounds in the algorithm. Moreover, it is important to note that the algorithm keeps this action pair $(i, j)$ for at least one additional round following the elimination of the optimal pair as it is identified as a NE.

On the other hand, for the action pairs in $S_{A_2}$, we assume that $ \hat{\Delta}_t < \frac{\lambda}{2}$, which implies that a suboptimal action pair remains in the game until the final round. This scenario emphasizes the regret contribution when the estimated suboptimality gap $\hat{\Delta}_t$ becomes sufficiently small, a case represented by the action pairs in $S_{A_2}$. Consequently, these suboptimal action pairs contribute to the overall regret and the analysis accounts for their effect to establish accurate performance guarantees.

We observe that the optimal action pair is eliminated by $(i,j)$ in some round $t_*$ where $t_*$ ranges from 0 to $t_{ij}$. Once this elimination occurs, the action pair action pair $(i,j)$ must remain active for at least one additional round, since it is selected as a misidentified NE. Hence, the algorithm necessarily keeps $(i,j)$  until at least round $t_*+1$. Furthermore, because $t_* \leq t_{ij}$, probability of keeping a suboptimal action pair $(i,j)$  in round $t_*+1$ can be bounded by $\frac{16 \sigma^2}{\hat{\Delta}_{t_{ij}+1}^2 T}$. Thus, we can write the regret contribution of this case as
\begin{align}
    \sum_{(i,j) \in S_{A}} \Delta_{ij}^* T \frac{16 \sigma^2}{\hat{\Delta}_{t_{ij}+1}^2 T} & = \sum_{(i,j) \in S_{A}} \Delta_{ij}^* T \frac{64 \sigma^2}{\hat{\Delta}_{t_{ij}}^2 T} \\
    & \leq \sum_{(i,j) \in S_{A_1}} \Delta_{ij}^*  \frac{512 \sigma^2}{\Delta_{ij}^2} + \sum_{(i,j) \in S_{A_2}} \Delta_{ij}^*  \frac{512 \sigma^2}{\lambda^2}
\label{eq:regret_contribution_case3}
\end{align}
where $\hat{\Delta}_{t+1} = \frac{\hat{\Delta}_t}{2}$ and we apply \eqref{ineq:Delta_t} to get \eqref{eq:regret_contribution_case3}. We consider the action pairs in $S_{A_2}$ for this case, as this suboptimal action pair is identified as a NE. This is necessary because we need to account for regret from the commit phase in Algorithm \ref{alg:etc_with_action_elimination} and the round $t_{ij}+1$ might be the round when the commit phase started.

\begin{case}
    A suboptimal action pair $(i, j)$ from the set $S_{A_2}$ is selected as the unique action pair in $S_{t_{ij}}$ for remaining rounds.
\end{case}
This implies that a suboptimal action pair is played during the commit phase of Algorithm \ref{alg:etc_with_action_elimination} up to step $T$. Thus, we need to account for an additional regret contribution term given by a trivial upper bound $\max_{(i,j) \in S_{A_2} } \Delta_{ij}^* T$. 

This regret contribution accounts for scenarios where the algorithm selects suboptimal action pair as the pure NE. In other words, if the selected action pair is suboptimal, the algorithm will continue to play with this action pair for the remaining rounds. Since the action pair selection persists up to time step $T$, resulting in an additional term in the regret bound. 

Lemma \ref{lemma:relationbetw_delta*_and_delta} ensures that the regret does not grow excessively, and this relationship allows us to analyze the regret in terms of the defined action sets. As a result, if we sum these regret contributions as mentioned, we can conclude the proof.
\bbox

\section{Proof of Theorem \ref{thm:regretbound_nonuniformexploration}}
\label{section:proof_of_thm_nonuniform_exp}
We follow exactly the same steps as the proof of Theorem \ref{thm:regretbound_actionpairelimination}. We update the threshold parameter as $\lambda \geq 4 \sigma \sqrt[4]{e/T}$, thus we have two subsets of action pairs such that $S_{A_1} = \{ (i,j) \in S_A: \Delta_{ij} > \lambda \}$ and $S_{A_2} = \{ (i,j) \in S_A: 0 < \Delta_{ij} \leq \lambda \}$. We can use the same probability of keeping an action pair as in \eqref{eq:prob_of_keep_an_action_pair} since the algorithm each action pair at least $k_t$ times. Then, if we keep the same events from the proof of Theorem \ref{thm:regretbound_actionpairelimination}, Appendix \ref{section:appendix_proof_thm5_1}, we have the following: 
\begin{align}
    \mathbb{P}(E_1 \cap E_2) & \leq \min \{ \mathbb{P}(E_1), \mathbb{P}(E_2) \} \\ & \leq \min \left \{ \exp{\left( - \frac{ (\Delta_{ij}^{\text{max}} - \varepsilon_t)^2 }{2 ( \frac{\sigma^2}{k_{i'j,t}} + \frac{\sigma^2}{k_{ij,t}})} \right) }, \exp{\left( - \frac{ (\Delta_{ij}^{\text{min}} - \varepsilon_t)^2 }{2 ( \frac{\sigma^2}{k_{ij',t}} + \frac{\sigma^2}{k_{ij,t}})} \right) }  \right \} \\
    & \leq \min \left \{ \exp{\left( - \frac{  (\Delta_{ij}^{\text{max}} - \varepsilon_t)^2 }{2 \left( \frac{\sigma^2}{ (k_t + k_t\hat{P}_{i'j,t})} + \frac{\sigma^2}{(k_t + k_t \hat{P}_{ij,t})} \right)} \right) }, \exp{\left( - \frac{ (\Delta_{ij}^{\text{min}} - \varepsilon_t)^2 }{2 \left( \frac{\sigma^2}{ (k_t + k_t\hat{P}_{ij',t})} + \frac{\sigma^2}{(k_t + k_t \hat{P}_{ij,t})} \right)} \right) }  \right \} \\
    & \leq \min \left \{ \exp{\left( - \frac{ (\Delta_{ij}^{\text{max}} - \varepsilon_t)^2 }{2 \sigma^2 \left( \frac{(\hat{P}_{i'j,t} + \hat{P}_{ij,t} + 2)}{ k_t (1 +\hat{P}_{i'j,t}) (1 + \hat{P}_{ij,t})}  \right)} \right) }, \exp{\left( - \frac{ (\Delta_{ij}^{\text{min}} - \varepsilon_t)^2 }{2 \sigma^2 \left( \frac{(\hat{P}_{ij',t} + \hat{P}_{ij,t} + 2)}{ k_t (1 +\hat{P}_{ij',t}) (1 + \hat{P}_{ij,t})}  \right)} \right) }  \right \} \\
    & \leq \min \left \{ \exp{\left( - \frac{ k_t (\Delta_{ij}^{\text{max}} - \varepsilon_t)^2 }{4 \sigma^2} \right) }, \exp{\left( - \frac{ k_t (\Delta_{ij}^{\text{min}} - \varepsilon_t)^2 }{4 \sigma^2} \right) }  \right \} \label{ineq:pij's_to_2} \\
    & \leq \frac{16 \sigma^2}{\hat{\Delta}_t^2 T}.
\end{align}
Since the terms $\frac{(\hat{P}_{i'j,t} + \hat{P}_{ij,t} + 2) }{ (1 +\hat{P}_{i'j,t}) (1 + \hat{P}_{ij,t})}$ and $\frac{(\hat{P}_{ij',t} + \hat{P}_{ij,t} + 2) }{ (1 +\hat{P}_{ij',t}) (1 + \hat{P}_{ij,t})}$ are obviously less than 2 since $0 \leq \hat{P}_{ij,t} \leq 1$ for any $(i,j) \in S_A$, the inequality \eqref{ineq:pij's_to_2} follows. Then, it leads to the same probability of keeping an action pair as in the proof of Theorem \ref{thm:regretbound_actionpairelimination} and also to the same results for Case 1, Case 3. 

On the other hand, since we explore each action pair non-uniformly, we need to update Case 2 by using 
\begin{align}
    k_{t_{ij}} = \left \lceil \frac{16 \sigma^2}{\hat{\Delta}_{t_{ij}}^2} \ln{\left(\frac{\hat{\Delta}_{t_{ij}}^2 T}{16 \sigma^2} \right)} \right \rceil + \left \lceil \frac{16 \sigma^2 \hat{P}_{ij,t_{ij}}}{\hat{\Delta}_{t_{ij}}^2} \ln{\left(\frac{\hat{\Delta}_{t_{ij}}^2 T}{16 \sigma^2} \right)} \right \rceil \leq 2 \ \left \lceil \frac{256 \sigma^2}{\Delta_{ij}^2} \ln{\left(\frac{\Delta_{ij}^2 T}{256 \sigma^2}\right)} \right \rceil
\end{align}
since $\hat{P}_{ij,t_{ij}} \leq 1$ for all $i,j$. 
Then, the regret contribution becomes
\begin{align}
    \sum_{(i,j) \in S_{A_1}} \Delta_{ij}^* \ 2 \ \left \lceil \frac{256 \sigma^2}{\Delta_{ij}^2} \ln{\left(\frac{\Delta_{ij}^2 T}{256 \sigma^2}\right)} \right \rceil 
    & \leq \sum_{(i,j) \in S_{A_1}} \Delta_{ij}^* + \Delta_{ij}^* \frac{512 \sigma^2}{\Delta_{ij}^2} \ln{\left(\frac{\Delta_{ij}^2 T}{256 \sigma^2}\right)} 
\end{align}
Then, it concludes the proof.

\begin{theorem}
    The external regret of Algorithm \ref{alg:etc_with_nonuniform_exploration} is bounded by 
\begin{align}
    R_T  \leq \sum_{(i,j) \in S_{A_1}} \Delta_{ij}  + \frac{768 \sigma^2}{\Delta_{ij}} + \frac{512 \sigma^2} {\Delta_{ij}} \ln{\left(\frac{\Delta_{ij}^2 T}{256 \sigma^2}\right)} \notag + \sum_{(i,j) \in S_{A_2}} \left( \frac{512 \sigma^2}{\lambda} +  \lambda T \right).
\end{align}
\end{theorem}
Since we have $\Delta_{ij}^* \leq \Delta_{ij}$ from Lemma \ref{lemma:relationbetw_delta*_and_delta}, it concludes the proof.

\section{Experimental Details}
\label{section:appendix_experiments}
In this section, we present additional experiments to compare the performances of the algorithms  based on their theoretical bounds and cumulative regret. To calculate the cumulative regret, we use the absolute difference between the value of the game and the payoff of the action played as the regret might otherwise be negative.

\begin{figure}[ht!]
    \centering
    \includegraphics[width=0.75\linewidth]{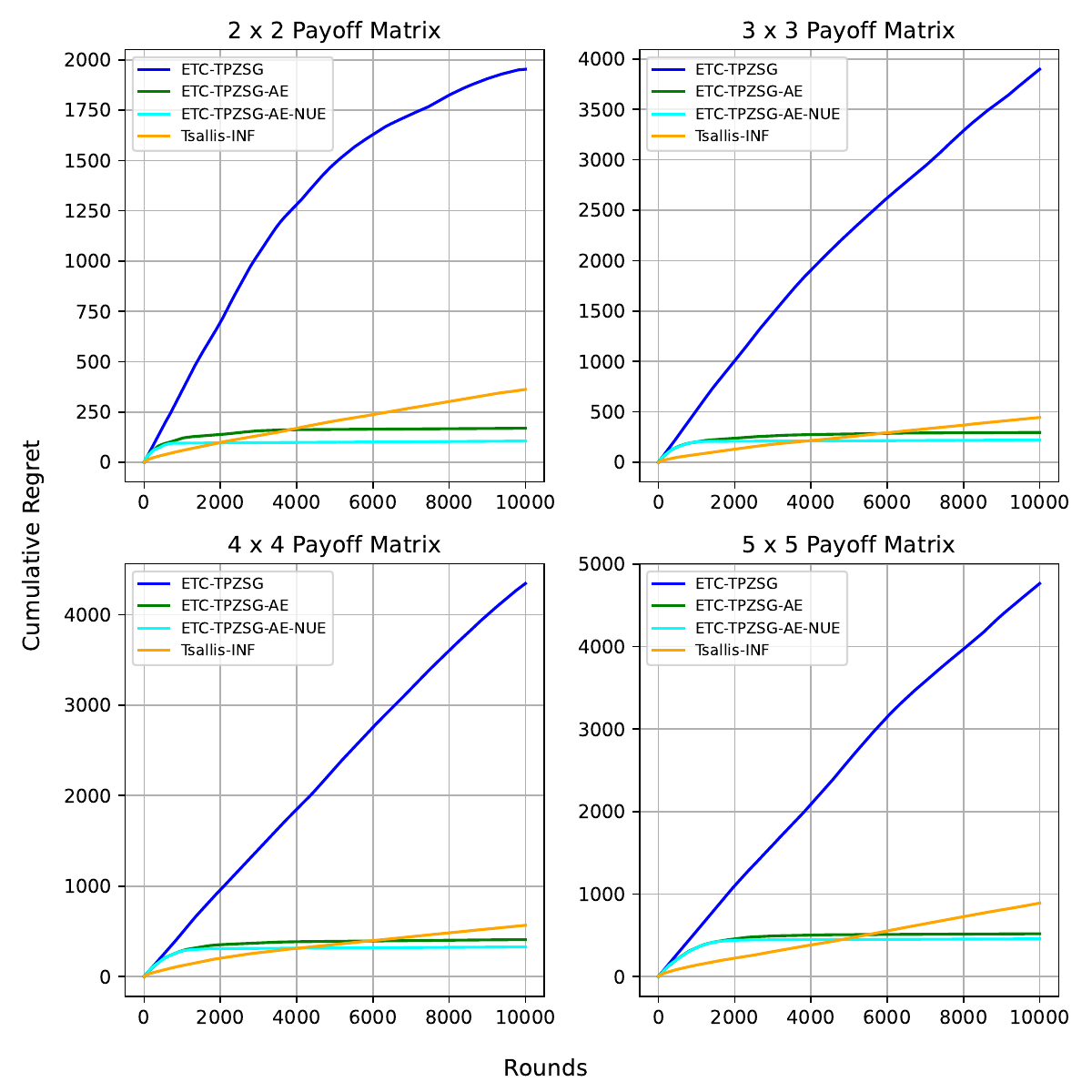}
    \caption{The cumulative regrets of the algorithms for different game matrices with large $\Delta$}
    \label{fig:algo_comp_differentgames_largedelta}
\end{figure}

\begin{figure}[ht!]
    \centering
    \includegraphics[width=0.75\linewidth]{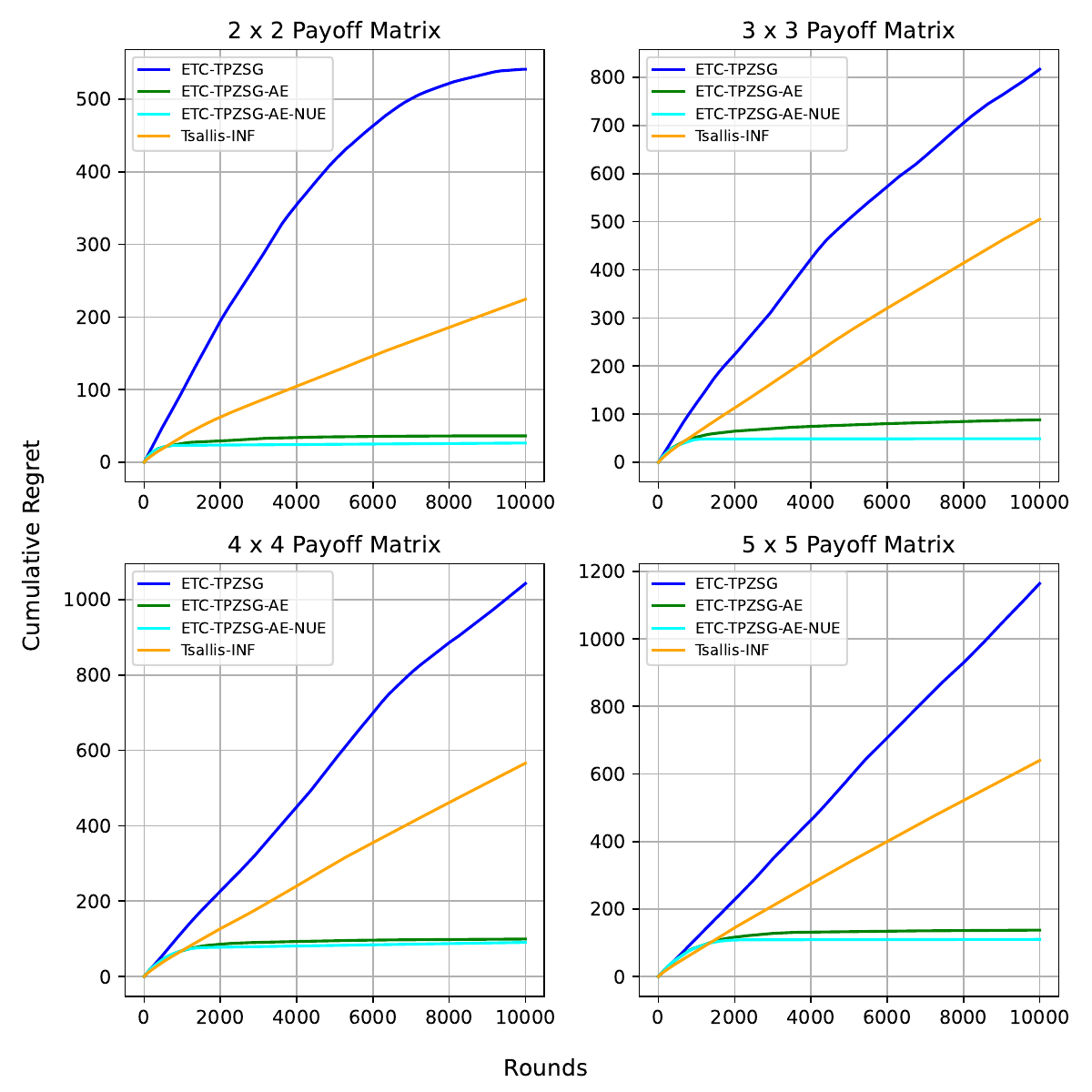}
    \caption{The cumulative regrets of the algorithms for different game matrices with small $\Delta$}
    \label{fig:algo_comp_differentgames_smalldelta}
\end{figure}

In Figures \ref{fig:cum_regrets_of_algorithms_large_delta} and \ref{fig:cum_regrets_of_algorithms_small_delta} , we compare the performance of the \ETC{}, \ETCAE{}, \ETCNUE{} and Tsallis-INF \cite{ito2025instance} algorithms. In each run, the exploration time $k$ for each action pair in \ETC{} is randomly selected from a predefined list ranging from 100 to 2500 in increments of 25. The total number of rounds is set to $T = 10^4$. On the other hand, each run uses a new game matrix with a unique pure NE which is generated using Gaussian payoffs with mean zero, standard deviation $\sigma$ selected from the list $[ 0.1, 0.2 ]$ for Figure \ref{fig:cum_regrets_of_algorithms_small_delta} and from the list $[ 0.25, 0.5, 0.75, 1 ]$ for Figure \ref{fig:cum_regrets_of_algorithms_large_delta}. Here, $\sigma$ denotes the variance factor of Gaussian payoffs and setting it to a small value leads to smaller $\Delta$ values. Furthermore, varying $\sigma$ not only changes the payoff distribution but also affects the exploration time and the tolerance parameter $\varepsilon$ used in the elimination phase of \ETCAE{}. The cumulative regret is averaged over $10^2$ simulation runs. Furthermore, Figures \ref{fig:algo_comp_differentgames_largedelta} and \ref{fig:algo_comp_differentgames_smalldelta} present the comparison of algorithms based on different game matrices. For large $\Delta$, although the Tsallis-INF algorithm performs better in the first $10^3$ rounds, over time, our adaptive elimination algorithms reach lower cumulative regret.

As observed, the \ETCAE{} algorithm consistently outperforms \ETC{}, demonstrating the effectiveness of its elimination strategy in reducing cumulative regret, and \ETCNUE{} outperforms even for larger action sets. 
Increasing the number of actions for players makes the identification of the true NE more challenging, as reflected in the performance of \ETC{} as shown in Figures \ref{fig:algo_comp_differentgames_largedelta} and \ref{fig:algo_comp_differentgames_smalldelta}. On the other hand, when $\Delta$ is smaller, Tsallis-INF \cite{ito2025instance} also has difficulty in identifying pure NE correctly, as seen in Figure \ref{fig:algo_comp_differentgames_smalldelta}. However, Algorithms \ref{alg:etc_with_action_elimination} and \ref{alg:etc_with_nonuniform_exploration} perform well even in these cases, as they incorporate pure NE properties during the elimination phase. In particular, ETC-based algorithms offer the advantage of conceptual and implementational simplicity.

\end{document}